\documentclass{bmvc2k}
\usepackage{lipsum}                
\usepackage{xargs}                     
\usepackage{graphicx}
\usepackage{tikz}
\usepackage{comment}
\usepackage{wrapfig}
\usepackage{amsmath,amssymb} %
\usepackage{color}
\usepackage{booktabs}
\usepackage{multirow}
\usepackage{dblfloatfix}
\usepackage{enumitem}
\usepackage[ruled,vlined]{algorithm2e}
\usepackage[accsupp]{axessibility}  %

\usepackage[capitalize]{cleveref}
\usepackage{amsfonts,bm}

\usepackage{grffile}

\title{Disentangling Content and Motion for Text-Based Neural Video Manipulation}

\addauthor{Levent Karacan}{levent.karacan@iste.edu.tr}{1}
\addauthor{Tolga Kerimoğlu, İsmail İnan}{{tolga.kerimoglu, ismail.inan}@boun.edu.tr}{2}
\addauthor{Tolga Birdal}{tbirdal@imperial.ac.uk}{3}
\addauthor{Erkut Erdem}{erkut@cs.hacettepe.edu.tr}{4}
\addauthor{Aykut Erdem}{aerdem@ku.edu.tr}{5}

\addinstitution{
 Computer Engineering Department\\
 Iskenderun Technical University
}
\addinstitution{
 Computer Engineering Department\\
 Boğaziçi University
}
\addinstitution{
 Department of Computing\\
 Imperial College London
}
\addinstitution{
 Computer Engineering Department\\
 Hacettepe University
}
\addinstitution{
 Computer Engineering Department\\
 Koc University
}

\runninghead{Karacan \etal}{DiCoMoGAN: Text-Based Neural Video Manipulation}

\def\eg{\emph{e.g}\bmvaOneDot}

\def\etal{\emph{et al}\bmvaOneDot}
\def\eqref#1{equation~\ref{#1}}
\def\1{\bm{1}}

\newcommand{\test}{\mathcal{D_{\mathrm{test}}}}

\def\gX{{\mathcal{X}}}
\def\gY{{\mathcal{Y}}}

\def\sR{{\mathbb{R}}}

\newcommand{\E}{\mathbb{E}}

\newcommand{\R}{\mathbb{R}}

\newcommand{\KL}{D_{\mathrm{KL}}}

\DeclareMathAlphabet{\mathsfit}{\encodingdefault}{\sfdefault}{m}{sl}
\SetMathAlphabet{\mathsfit}{bold}{\encodingdefault}{\sfdefault}{bx}{n}

\newcommand{\x}{\mathbf{x}}
\newcommand{\f}{\mathbf{f}}
\newcommand{\z}{\mathbf{z}}

\newcommand{\Man}{\mathcal{M}}

\newcommand{\ODEpars}{\bm{\theta}}

\newcommand{\zdyn}{\mathbf{z}^{\mathrm{dyn}}}
\newcommand{\zstatic}{\mathbf{z}^{\mathrm{ST}}}

\newcommand{\fnode}{{f}_{\mathrm{ODE}}}
\newcommand{\RepNet}{\mathrm{RepNet}}
\newcommand{\TraNet}{\mathrm{TraNet}}
\newcommand{\loss}{\mathcal{L}}

\newcommand{\w}{\mathbf{w}}

\newcommand{\y}{\mathbf{y}}

\newcommand{\Pro}{\mathbb{P}}

\newtheorem{dfn}{Definition}

\newcommand{\tolga}[1]{{\textcolor{black}{#1}}}

\newcommand{\ie}{\textit{i}.\textit{e}.}
\newcommand{\cf}{\textit{c}.\textit{f}.}

\renewcommand{\paragraph}[1]{{\vspace{1mm}\noindent \bf #1}.}
\newcommand{\paragraphnoper}[1]{{\vspace{1mm}\noindent \bf #1}}

\usepackage{cleveref}

\Crefname{assumption}{\textbf{H}\hspace{-3pt}}{\textbf{H}\hspace{-3pt}}
\crefname{algorithm}{\text{Alg.}}{\text{Alg.}}
\crefname{assumption}{\textbf{H}}{\textbf{H}}
\crefname{equation}{\text{Eq}}{\text{Eq}}
\crefname{definition}{\text{Dfn.}}{\text{Dfn.}}
\crefname{lemma}{\text{Lemma}}{\text{Lemma}}
\crefname{dfn}{\text{Dfn.}}{\text{Dfn.}}
\crefname{thm}{\text{Thm.}}{\text{Thm.}}
\crefname{tab}{\text{Tab.}}{\text{Tab.}}
\crefname{fig}{\text{Fig.}}{\text{Fig.}}
\crefname{table}{\text{Tab.}}{\text{Tab.}}
\crefname{figure}{\text{Fig.}}{\text{Fig.}}
\crefname{section}{\text{Sec.}}{\text{Sec.}}

\crefname{section}{Sec.}{Secs.}
\Crefname{section}{Section}{Sections}
\Crefname{table}{Table}{Tables}
\crefname{table}{Tab.}{Tabs.}

\begin{document}

\maketitle
\vspace{-4mm}
\begin{abstract}
Giving machines the ability to imagine possible new objects or scenes from linguistic descriptions and produce their realistic renderings is arguably one of the most challenging problems in computer vision. Recent advances in deep generative models have led to new approaches that give promising results towards this goal. In this paper, we introduce a new method called DiCoMoGAN for manipulating videos with natural language, aiming to perform local and semantic edits on a video clip to alter the appearances of an object of interest. Our GAN architecture allows for better utilization of multiple observations by \textbf{di}sentangling \textbf{co}ntent and \textbf{mo}tion to enable controllable semantic edits. To this end, we introduce two tightly coupled networks: (i) a \emph{representation network} for constructing a concise understanding of motion dynamics and temporally invariant content, and (ii) a \emph{translation network} that exploits the extracted latent content representation to actuate the manipulation according to the target description. Our qualitative and quantitative evaluations demonstrate that DiCoMoGAN significantly outperforms existing frame-based methods, producing temporally coherent and semantically more meaningful results.
\end{abstract}

\section{Introduction}
\label{sec:intro}

Making desired edits on an image or video using tools like Adobe Photoshop, Adobe Premiere Pro and Apple Final Cut Pro is quite challenging and requires extensive training and experience. Thanks to the proliferation of deep learning, some user-friendly solutions are proposed for editing images~\cite{li2020manigan}. Yet, democratizing the video editing process to improve accessibility and empower the non-experts still requires rethinking modern architectures. 

\begin{figure}[!t]
\centering
\includegraphics[width=0.9\linewidth]{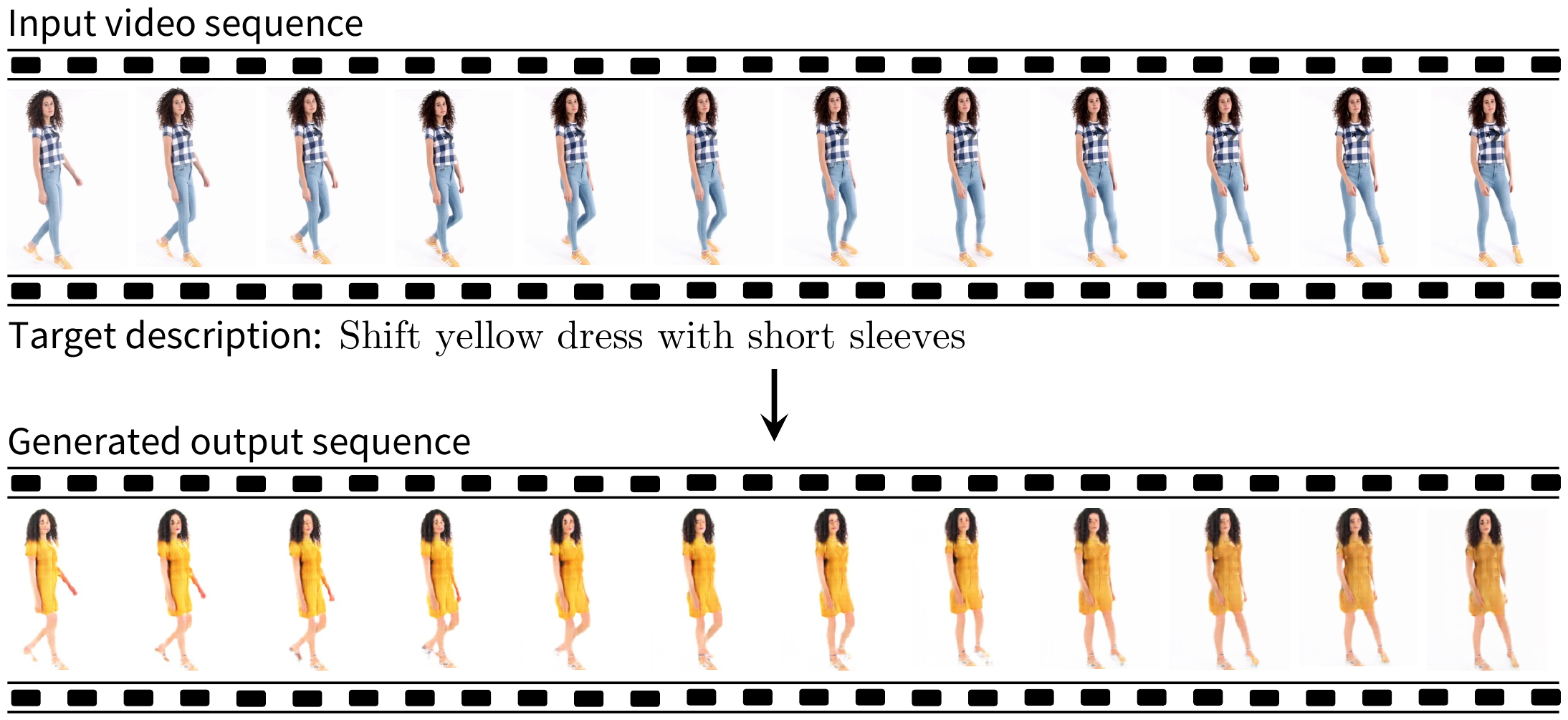}

\vspace{-2mm}
\caption{\textbf{Text-based video manipulation.} Given a video sequence and a target \mbox{description}, our DiCoMoGAN model generates a temporally coherent output sequence, carrying out the necessary structural changes while preserving the attributes not referred in the text (\eg~hair style, identity), and does this without any extra guidance like semantic layout information.\vspace{-4mm}
}
\label{fig:videomanipulation}
\end{figure}

Towards this end, we set off to ask: \emph{Can we learn to semantically manipulate videos through natural language descriptions in a temporally consistent way?} (\cf~\cref{fig:videomanipulation})
The existing literature approaches this problem on a frame-by-frame basis applying minimal necessary modifications specified by the input text, disjointly to the each and every input frame. Almost all of the image editing methods use encoder-decoder architectures~\cite{reed2016generative,dong2017semantic,nam2018text,li2020manigan,li2020lightweight,Xia2021TediGAN,Patashnik2021StyleCLIP} and employ adversarial learning strategies~\cite{isola2017image,karras2019style} to provide the agreement between the resulting images and the target text and to generate photo-realistic outcomes. However, performing language-driven edits on videos requires models to not only understand the frame content but also be aware of the global video context and its temporal unidirectionality. Moreover, a harmonious editing demands the target descriptions to be related not only to static frames but to the entire video to achieve \emph{good gestalt}. 

To achieve all these, we propose a new data-driven text based video manipulation model called \textbf{DiCoMoGAN}. The key to our approach is a unified network model consisting of a \emph{representation network} (RepNet) and a \emph{translation network} (TraNet), which jointly learn to disentangle video content and motion dynamics and to perform the text-specified edits on a given video sequence. Under the assumption that textual description is strongly related to appearance, we create a structured latent space composed of \emph{text relevant}, \emph{text irrelevant} and \emph{dynamic} subspaces (\cf~\cref{fig:systemoverview}). To ensure the former, we steer the latent subspace to be shared between global video descriptor and the text features, encoded by CLIP~\cite{Patashnik2021StyleCLIP}. We then use the features from this structured latent space along with text features to condition \emph{multi-feature modulation} (MFMOD) blocks. We train this integrated architecture via a multi-task loss function in an end to end manner to encode scene specific transformations effectively while capturing the relationships between the spatiotemporal data and the text input. 
Our experiments on the standard 3D Shapes benchmark~\cite{3dshapes18} as well as on our new dataset \emph{Fashion Videos} demonstrate that \mbox{DiCoMoGAN} can produce high quality, temporally consistent videos faithfully reflecting the \emph{intentions} stated in the target descriptions. In summary, our contributions are as follows: (1) Our representation network, RepNet, implements a neural architecture that explicitly enforces the separation of static and dynamic features via a set-based $\beta$-VAE model~\cite{Higgins2017betaVAELB} equipped with a Latent ODE~\cite{rubanova2019latent}. (2) Our translation network, TraNet, follows an encoder-decoder architecture which is guided by the representation network through a novel multi-feature modulation block called \mbox{MFMOD} where the residual activation maps are modulated based on both the given textual description and the disentangled content code. (3) To test the capabilities of our model in a more realistic setting, we collect a new dataset containing \emph{Fashion Videos} with the related textual descriptions.

\section{Related Work}
\label{sec:relatedwork}
\paragraph{Disentangled representations} 
The aim of unsupervised disentangled representation learning is to discover underlying factors of variation in a training data, in which each dimension encodes a unique and semantically meaningful aspect of the data~\cite{bengio2013representation}. To this end, most of the existing approaches are based on VAEs~\cite{kingma2014auto,rezende2014stochastic} with slight modifications in the VAE objective, such as $\beta$-VAE~\cite{Higgins2017betaVAELB}, FactorVAE~\cite{kim2018disentangling}. Similarly, some prior work tweak the objective of GANs~\cite{goodfellow2014generative} to achieve disentanglement, \eg InfoGAN~\cite{Chen16infoGAN}, IB-GAN~\cite{jeon2021ibgan}. Locatello \etal ~\cite{locatello2019challenging} showed that unsupervised learning of disentangled generative factors can not be achieved without strong inductive biases on both the models and the data, which can be alleviated using weak supervision or a few labeled training samples~\cite{trauble2021disentangled}. Key to the success of our model, in our work we especially focus on disentangling motion and content, which has been previously studied in a fully unsupervised setting~\cite{Denton2017}, or using action/attribute labels~\cite{he2018probabilistic}. But we instead utilize natural language descriptions as weak supervision.

\paragraph{VAE-GAN hybrids} GANs are superior to VAEs in terms of visual quality, but VAEs provide better disentangled representations. There is a line of research that explores combining VAE and GAN frameworks, ALI~\cite{dumoulin2017adversarially}, BiGAN~\cite{donahue2017adversarial}, IntroVAE~\cite{IntroVAE}, to name a few. The promise of these so-called hybrid approaches is to combine the advantages of both models, while providing a much stable training and improved diversity in the generated samples.
\raggedbottom

\paragraph{Video synthesis} 
Video synthesis aims at
generating temporally coherent video clips either from scratch~\cite{Mathieu2016,Finn2016,Xue2016,Liang2017,Denton2017,Weissenborn2020iclr}, or according to a single image or a short sequence of images~\cite{Vondrick2016,Saito2017,Tulyakov2018,denton2018icml,akan2021iccv}. Motivated by these works, there are also some efforts to allow the users to control the video generation process by introducing natural text~\cite{Marwah2017,Pan2017,Li2018,Balaji2019} or spoken language~\cite{Zhou2019,Das2020,Thies2020,yang2020face,Chen2021} as an additional input.

\paragraph{Language based image manipulation} 
In text-to-image synthesis, the goal is to generate an image with a natural language description~\cite{reed2016generative,zhang2017txt2im,xu2018txt2im,zhu2019txt2im}. On the other hand, semantic image manipulation aka language guided image editing models~\cite{dong2017semantic,nam2018text,li2020manigan,li2020lightweight} aim at modifying a source image according to a given textual description summarizing the desired object characteristics. SISGAN~\cite{dong2017semantic} involves a text-conditioned encoder-decoder architecture. TAGAN~\cite{nam2018text} learns to disentangle different semantic attributes of the target object during training by considering a text-adaptive discriminator. ManiGAN~\cite{li2020manigan} and LightweightGAN~\cite{li2020lightweight}, on the other hand, utilize text-image combination modules, which are used to match semantic attributes with certain words in the given descriptions, along with explicit word-level discriminators to improve the quality of the results. The recently proposed TediGAN~\cite{Xia2021TediGAN}, Latent Transformer~\cite{Yao_2021_ICCV},  and StyleCLIP~\cite{Patashnik2021StyleCLIP} models are also capable of performing language-driven edits on a given image, but they all require a StyleGAN model pre-trained for a specific domain (\eg faces), which is hard to train for less structured domains like full body images. 
Recently, Jiang et al.~\cite{JiangICCV2021} propose a new language-guided editing model specifically designed for performing global edits such as changing brightness or color tone of an image.

\paragraph{Language based video manipulation}
\tolga{Our task of video-editing using natural language descriptions is a relatively new one. There are two studies worth mentioning which are concurrent to this work: (i)~\cite{m3l} tackles a similar problem by proposing a transformer-based architecture, but they did not make their implementation freely available; (ii)~\cite{charm} presents a StyleGAN3-based video-editing framework, but, it only considers manipulations based on a single attribute. The latter belongs to the family of GAN-inversion based~methods which uses the latent space of a pretrained StyleGAN model for editing purposes, where the existing methods focus on distinct domains such as aligned faces, as the style-based generators do not work well on unstructured datasets. Note that while inversion constitutes a sensible research direction for GANs, inverting diffusion models without significant distortion remains a challenge. Exploring these directions is future work. Nonetheless, our approach and concept of disentangled video editing can be used regardless of the backbone architectures.}

\section{DiCoMoGAN}\label{sec:method}
\begin{figure*}[!t]
\centering
\includegraphics[width=0.9\linewidth, ]{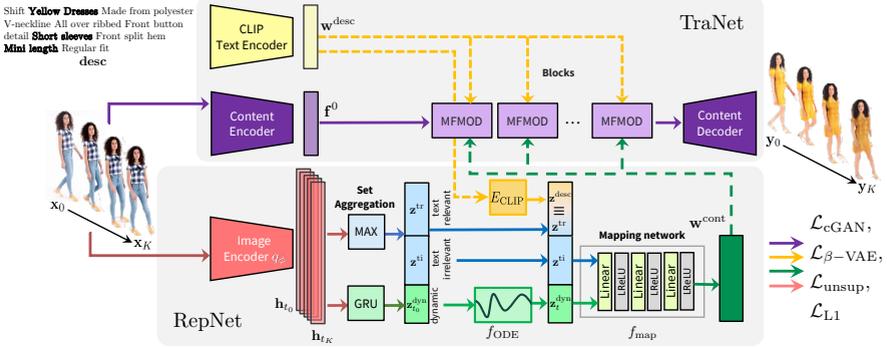}
\caption{\textbf{Schematic illustration of our DiCoMoGAN model.} DiCoMoGAN consists of a representation network (RepNet) and a translation network (TraNet) trained in a harmonious manner. While the former aims to disentangle motion and content by a set-based formulation combining VAEs, latent ODEs and set operations, the latter takes advantage of the extracted latent features to better guide the manipulation by employing a conditional normalization method which we call multi-feature modulation (MFMOD).\vspace{-5mm}}
\label{fig:systemoverview}
\end{figure*}

\paragraph{Problem setting}
\tolga{We consider the problem of manipulating a given input video according to a provided textual description.} Inputs to our text-based video manipulation approach, called DiCoMoGAN, are a short video clip of a single object and a target text description summarizing the object's new look. We represent the source video as an image sequence denoted by $X = (\mathbf{x}_i\in\mathbb{R}^{3\times H\times W})_{i=1}^N$,  with $i$ being the frame index and $N$ indicating the total number of frames. Our goal is to perform seamless and semantically meaningful edits on each video frame $\mathbf{x}_i$ to reflect what is being described in the target text $\mathbf{desc}$, and accordingly generate an output sequence $Y = (\mathbf{y}_i)_{i=1}^N$ of the same spatial dimensions as the input -- with the desired look. Our model carries this out by:
\begin{enumerate}[itemsep=0pt,topsep=0pt,leftmargin=*]
    \item a \textbf{representation network} to learn a disentangled latent space in which static and dynamic semantic scene characteristics are encoded independently,
    \item a \textbf{translation network} capable of transferring the target look stated in the textual description to the source video in a truthful and temporally coherent manner, and
    \item unifying (1) and (2) with a \textbf{combined neural architecture} where the two networks are trained simultaneously in an end-to-end manner. %
\end{enumerate}
In what follows, we describe DiCoMoGAN in detail, following the structure shown in~\cref{fig:systemoverview}.\vspace{-2mm}

\subsection{Network Architecture}
\paragraph{Representation network (RepNet)} 
Unlike prior work~\cite{dong2017semantic,nam2018text} focusing on language-driven image edits, our aim is to perform edits on short video clips. Thus, in our formulation, capturing the intrinsic characteristics of the scene and the object depicted in the source video plays a key role. As a remedy, we design RepNet for the purpose of extracting a disentangled representation of the input video from the complementary video frames $\{\mathbf{x}_i\}$. To this end, we employ a $\beta$-VAE architecture~\cite{Higgins2017betaVAELB} enriched with a Latent ODE~\cite{rubanova2019latent} to encode an input video $X$ in a latent space. In particular, we split the latent space into two parts as static and dynamic: $\z = \left[\begin{array}{@{}c@{\;\;}c@{}}
 \z^{\mathrm{ST}} & \zdyn\end{array}\right]$. Static latent codes $\zstatic$ do not change across consecutive frames and encode properties like object color and identity, etc. Dynamic codes $\zdyn$ are steered by the Latent ODE and encode characteristics that smoothly vary across frames like pose, orientation. \tolga{Such explicit architecture design coupled with respective loss functions (to be precised later) builds the appropriate inductive bias, encoding distinct features by certain dimensions of the latent space.}

The main part of RepNet is an image encoder network $q_\phi$ including CNN layers, a GRU module and a Neural ODE~\cite{chen2018neural} which is responsible for extracting dynamic latent codes. We assume that RepNet takes a set of frames $\{\x_j\}_{j=0}^K$ at times $\{t_j\}_{j=0}^K$, where $K (K<N)$ denotes the number of frames irregularly sampled from the input video clip. Its convolutional layers encode each observation individually to a feature map, resulting in the set $\{\mathbf{h}_j\}_{j=t_0}^{t_{K}}$. 

Inspired by~\cite{rempe2020caspr}, we disentangle the motion dynamics from appearance (content). To obtain the \emph{static}, \ie non-time varying latent codes, we first max-pool a subspace of those hidden features to get a permutation invariant representation $\hat{\mathbf{h}}^{\mathrm{ST}}$, which is then mapped to a static latent code $\zstatic$ through a linear layer. Note, $\zstatic$ is shared among all the input video frames and carries the global context. Dynamic codes are obtained by feeding the hidden features in the remaining subspace to a GRU module in reverse order with time gaps $\Delta t = t_i - t_{i-1}$ according to time stamps ($t_{K}>t_{K-1}>...>t_0$). GRU module produces a dynamic hidden feature $\hat{\mathbf{h}}^{\mathrm{dyn}}_{t_0}$ at $t_0$ using the update rule as given by:
\begin{equation}
    \hat{\mathbf{h}}^{\mathrm{dyn}}_{t_{i-1}} = \mathrm{GRU}(\hat{\mathbf{h}}^{\mathrm{dyn}}_{t}, \Delta t, \mathbf{h}_{t_{i-1}})\;,
\end{equation}
A linear layer then maps $\hat{\mathbf{h}}^{\mathrm{dyn}}_{t_0}$ to a dynamic latent code $\zdyn_{t_0}$. Once the dynamic latent code $\zdyn_{t_0}$ is calculated for $t_0$, a Neural ODE function $\fnode$ is learned to predict dynamic latent codes $\zdyn_{t}$ of the input video at all time stamps $t = t_0, t_1, ..., t_{K}$ using an ODE solver:
\begin{align}
    [\z_{t_{0}}^{\mathrm{dyn}} ,&\z_{t_{1}}^{\mathrm{dyn}},... \z_{t_{K}}^{\mathrm{dyn}}] = \mathrm{ODESolve}(\fnode, \z_{t_0}^{\mathrm{dyn}}, (t_0, t_1, ..., t_{K}))
\end{align}
The final latent code $\z_{t_i}$ for an input video at time step $t_i$ is built up by concatenating the calculated static and dynamic latent vectors as 
    $\z_{t_i} = \left[\begin{array}{@{}c@{\;\;}c@{}}
 \z^{\mathrm{ST}} & \zdyn_{t_i}
\end{array}\right]$.
From here on, we omit time subscripts whenever possible for notational convenience, \ie~$\z$ for $\z_{t_i}$.

\tolga{Our task requires learning to make local structural changes depending on the input text description like completely replacing an outfit with a new one. As such, text irrelevant regions must be preserved while performing required changes. Hence, we introduce a} modified $\beta$-VAE to learn to pass only text irrelevant codes to TraNet as condition. To this end, we split $\zstatic$ into \emph{text relevant} $\z^{\mathrm{tr}}$ and \emph{text irrelevant} $\z^{\mathrm{ti}}$ parts. To ensure better disentanglement in the latent space, we jointly let $\z^{\mathrm{tr}}$ live in the space of text features \ie~$\z^{desc}$, whose details will be precised in~\cref{sec:train}. Altogether, we write $\z^\prime = \left[\z^{\mathrm{tr}}\;\;\z^{\mathrm{ti}}\;\;\zdyn\right]$ %
representing the video frame at $t_i$. This representation aggregates information from multiple frames and is informed about the temporal dynamics. This cue is key in guiding TraNet in manipulating the source frames according to the target text.
In what follows, we pass the text irrelevant latent codes %
$\z^{\mathrm{cont}} = \left[\begin{array}{@{}c@{\;\;}c@{}}
 \z^{\mathrm{ti}} & \zdyn\end{array}\right]$ to TraNet as the content condition.

\paragraph{Translation network (TraNet)} As argued before, guiding the manipulation process based only on the target text is suboptimal since the textual description usually carries little information about which image regions to keep unchanged. Hence, we design TraNet as an encoder-decoder network with multiple conditioned residual blocks resembling a combination of pix2pixHD~\cite{wang2018high} and Semi-StyleGAN~\cite{nie2020semi}. The latent motion and static codes extracted from multiple frames help alleviating this by bringing additional conditioning. 

\begin{wrapfigure}{r}{0.4\textwidth}
\centering%
\includegraphics[width=\linewidth]{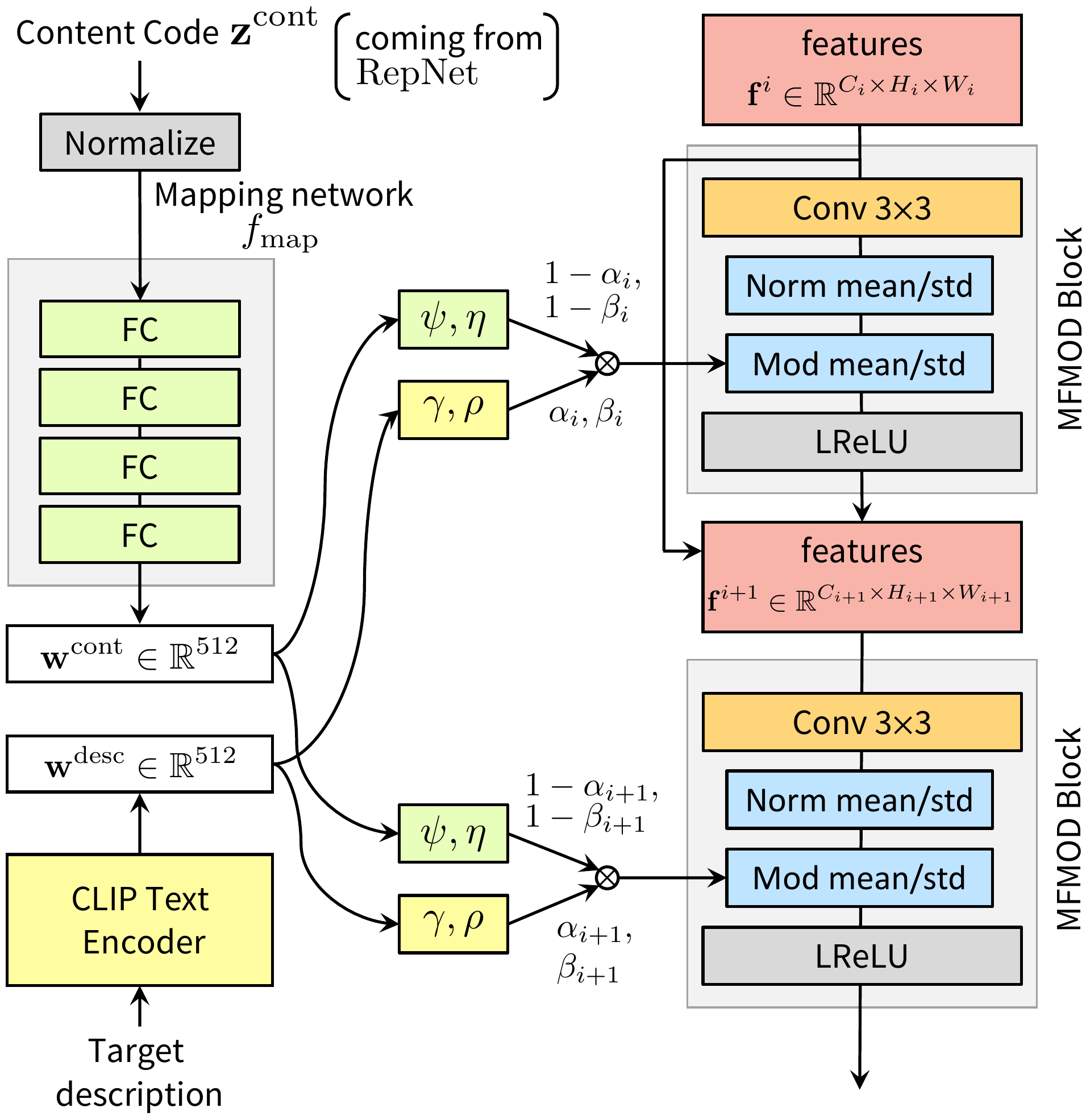}
\caption{\textbf{MFMOD block.} The proposed conditional normalization scheme modulates residual activation maps based on text and content code by learning optimal modulation parameters and blending weights.}%
\label{fig:mfmod}
\end{wrapfigure}

TraNet uses a special conditional normalization method, which we call \emph{multi-feature modulation} (MFMOD), that modulates the residual feature activation maps based on both text and content codes derived from RepNet. It learns optimum weights for each of these two conditions to perform feature modulation in an harmonious manner (\cf~\cref{fig:mfmod}). 

For a batch of $N$ samples, let the activation map before the $i^{th}$ MFMOD block be $\f^i\in\mathbb{R}^{C_i\times H_i\times W_i}$ where $C_i$ is the number of feature map channels and $H_i$ and $W_i$ are the spatial dimensions. The text condition $\w^{\mathrm{desc}}\in\sR^{512}$ is obtained by the text encoder of the pretrained CLIP  model~\cite{radford2021learning}. For the content latent $\z^{\mathrm{cont}}$, however, we employ a mapping network $f_{\mathrm{map}}$ (a shallow subnetwork composed of 4 fully connected layers) to map $\z^{\mathrm{cont}}$ to a higher-dimensional vector $\w^{\mathrm{cont}}\in\sR^{256}$. Our proposed normalization scheme can be interpreted as a special AdaIN operation~\cite{huang2017AdaIN} with an adaptive multi-feature blending before the activation feature modulation (\cf~\cref{fig:mfmod}). The normalized activation value at site $(n \in N, c \in C_i, y\in H_i, x \in W_i)$ is given by 
{\small
\begin{equation}
\centering
\begin{array}{ll}
\big(\alpha_{i}
\gamma^{i}_{c,y,x}(\w_{\mathrm{desc}})\,+\, (1-\alpha_i)\psi^{i}_{c,y,x}(\w_{\mathrm{cont}})\big)\frac{h^i_{n,c,y,x}-\mu^i_c}{\sigma^i_c} \,\,+\,\, \beta_i\rho^{i}_{c,y,x}(\w_{\mathrm{desc}}) \,\,+\,\, (1-\beta_i)\eta^{i}_{c,y,x}(\w_{\mathrm{cont}}) 
\end{array}
\end{equation}}
where $f^i_{n,c,y,x}$ is the preactivation,
$\mu^i_c$, $\sigma^i_c$ are the mean and the standard deviation of the activations in the channel $c$ given by:
{\small
\begin{equation}
\mu_{c}^{i} =\frac{1}{N H^{i} W^{i}} \sum_{n, y, x} f_{n, c, y, x}^{i}, \qquad \sigma_{c}^{i} =\sqrt{\frac{1}{N H^{i} W^{i}} \sum_{n, y, x}\left(\left(f_{n, c, y, x}^{i}\right)^{2}-\left(\mu_{c}^{i}\right)^{2}\right)},\nonumber
\end{equation}}
with $\gamma^{i}_{c,y,x}$, $\rho^{i}_{c,y,x}$, $\psi^{i}_{c,y,x}$, $\eta^{i}_{c,y,x}$ respectively denoting the learned modulation parameters for the description and content conditions $\w_{\mathrm{desc}}$ and $\w_{\mathrm{cont}}$. Note that the blending values $\alpha_i$ and $\beta_i$ are not fixed, but learned during the training phase.

After multi-conditional residual blocks, the last (conditioned) feature map is fed to the decoder, which consists of several convolutional transpose layers to upsample it to the original resolution to obtain the manipulated frame $\mathbf{y}$. In the decoder, we also apply instance normalization in all convolutional transpose layers except the last layer. We use ReLU activation in all convolutional and convolutional transpose layers in all parts of the network. Please refer to the supplementary material for details.

\subsection{Training}
\label{sec:train}
Our representation and translation networks, RepNet and TraNet, are trained jointly in an end to end manner, by minimizing a non-convex, multi-task loss:
\begin{align}
\loss  &= \loss_{\RepNet} + \lambda_{T}\loss_\TraNet, &
\loss_{\RepNet}&=\left(\loss_{\mathrm{rec}} + \loss_{\mathrm{rec}^\prime}\right)/2 - \beta(\loss_{\mathrm{KL}}^{\mathrm{ST}} + \loss_{\mathrm{KL}}^{\mathrm{dyn}})\\
& & \loss_\TraNet & = \min_{\TraNet}\bigg(\max_{\mathrm{Discr}}\loss_{\mathrm{cGAN}} + \lambda_{\mathrm{L1}}\loss_{\mathrm{L1}} + \lambda_{\mathrm{U}}\loss_{\mathrm{unsup}}\bigg)\nonumber
\end{align}
where $\lambda_{\mathrm{L1}}=1, \lambda_{\mathrm{U}}=0.5$, $\lambda_T=1$ are set empirically. We now define each of the loss terms.

\paragraph{Disentanglement losses  $\loss_{\mathrm{KL}}^{\mathrm{dyn}}$ and $\loss_{\mathrm{KL}}^{\mathrm{ST}}$}
We enforce the latent code $\z$ to disentangle latent factors of variation. This is measured by computing the KL-divergence individually for static and dynamic distributions:
{\small
\begin{equation}
\loss_{\mathrm{KL}}^{\mathrm{ST}}=\frac{1}{K}\sum_{t=0}^{K}\mathbb{E}_{q_{\phi}(\zstatic| \mathbf{x}_t)} D_{KL}(q_{\phi}(\zstatic | \mathbf{x}_t) \| p(\zstatic)), \quad \loss_{\mathrm{KL}}^{\mathrm{dyn}}=\mathbb{E}_{q_{\phi}\left(\zdyn_{t_0}| \mathbf{x}_{t_0}\right)} D_{K L}\left(q_{\phi}\left(\zdyn_{t_0} | \mathbf{x}_{t_0}) \| p(\zdyn_{t_0}\right)\right) %
    \nonumber
\end{equation}}%

\paragraph{Reconstruction losses $\loss_{\mathrm{rec}}$ and $\loss_{\mathrm{rec^\prime}}$}
During training $\z_{t_i}$ is passed to the image decoder $p_\theta$ to reconstruct the input video frame at time $t_i$ from its latent code. This reconstruction loss in our $\beta$-VAE objective reads:
\begin{equation}
\begin{aligned}
    \loss_{\mathrm{rec}}=\mathbb{E}_{q_{\phi}(\z| \mathbf{x})}
    \left[\log p_{\theta}(\mathbf{x} | \z)\right]\\
    \end{aligned}
\end{equation}
We also introduce an auxiliary text encoder $E_{\mathrm{CLIP}}$, whose output is aligned with text relevant code $\z^{\mathrm{tr}}$ creating a joint latent space. From the given textual description, text features $\textbf{w}^{\mathrm{desc}}$ are extracted by first using the text encoder $E_{\mathrm{CLIP}}$ of the off-the-shelf CLIP model~\cite{radford2021learning} and then feeding these CLIP embeddings to a series of linear projects to obtain a lower dimensional text representation $\textbf{z}^{\mathrm{desc}}$ that is the same dimension with that of $\textbf{z}^{\mathrm{tr}}$. We then define an additional reconstruction loss for learning to specify text relevant part of latent space as:
\begin{equation}
\begin{aligned}
    \loss_{\mathrm{rec}^\prime}=\mathbb{E}_{q_{\phi}(\z| \mathbf{x})}
    \left[\log p_{\theta}(\mathbf{x} | \mathbf{{z}'})\right]\;,
    \end{aligned}
\end{equation}
\tolga{This extra supervision enforces text relevant subspace of the latent code to be aligned with the CLIP space, improving the disentanglement ability of RepNet.}
Note that the image decoder $p_\theta$ and the auxiliary text encoder $E_{\mathrm{CLIP}}$ is not used in inference. 

\paragraph{GAN loss $\loss_{\mathrm{cGAN}}$}
The first cue for training TraNet comes from the conditional adversarial loss. %
We employ a discriminator network $\mathrm{Discr}$, which resembles the multi-scale PatchGAN discriminator~\cite{isola2017image,wang2018high}, with the only difference being the proposed MFMOD normalization block added after the last conv layer to improve conditioning. 

\paragraph{Perceptual loss $\loss_{\mathrm{L1}}$} To ensure the quality of the generated images, we employ a \emph{perceptual loss}~\cite{johnson2016perceptual} that minimizes the $L1$ distance between the feature maps of each input frame $\mathbf{x}_i$ and the manipulation result $\mathbf{y}_i$ extracted by a VGG-19 network~\cite{Simonyan15VGG} trained on ImageNet~\cite{deng2009imagenet}.
\begin{equation}
    \loss_{\mathrm{L1}} = \|\Phi_{VGG}(\mathbf{x})-\Phi_{VGG}(\mathbf{y})\|_1
\end{equation}
\paragraph{Unsupervised loss $\loss_{\mathrm{unsup}}$} 
Finally, to enforce consistency between latent codes of input frames ($x$) and their manipulated versions ($y$), we introduce an unsupervised loss defined as the $L_2$-distance between outputs from the $\beta$-VAE encoder of RepNet as:
\begin{equation}
    \loss_{\mathrm{unsup}} = \|q_{\phi}(\z|\mathbf{x}) - q_{\phi}(\z|\mathbf{y})\|_2
\end{equation}
We observe that while the contribution of this unsupervised loss to the final quality is only marginal, it helps to stabilize the training process.

\paragraph{Training details} We adopt a learning schedule on the image encoder of RepNet while back propagating the loss from TraNet. This is because at the beginning of training, content code from RepNet is incomplete, which hurts the training of TraNet. In particular, we gradually increase the learning rate from zero to a certain value along certain number of iterations.
We provide further details in the supplementary material.

\section{Experiments}
\label{sec:exp}
\paragraph{Datasets}
First, we use the \textbf{3D Shapes dataset}~\cite{3dshapes18} which is proposed for learning and assessing factors of variation from data. This dataset has 480K images of $64\times64$ resolution. There are $6$ ground truth independent latent factors. They are \textit{floor color}, \textit{wall color}, \textit{object color}, \textit{scale}, \textit{shape} and \textit{orientation}. For our purpose, we build simple text descriptions which covers object related latent factors \textit{object color}, \textit{scale} and \textit{shape}, \eg  \textit{``There is a big blue capsule.''}. To prevent scale ambiguity, we remove two elements of the scale factor which is of length $8$, originally. In that case, ``small'', ``medium'' and ``big'' in the descriptions correspond to the first two, middle two and the last two values, respectively. Moreover, we consider the orientation factor as a dynamic dimension taking $15$ different values. We have 19.2K train and 4.8K test videos with $15$ frames and simple text descriptions for each video. 

Second, to explore how well our model generalizes to more challenging datasets,  we collected a new video dataset, \textbf{Fashion Videos}, from an online shopping site, containing short video clips of individuals wearing different kinds of garments. Each clip includes full-body images of a single person moving around a scene, showing how the clothing looks from different angles. Moreover, the clip is endowed with a textual product description of the garment, detailing its visual features (color, material properties, and design details) as well as its category (\textit{dress}, \textit{jumpsuits}, \textit{trousers}, \textit{jumper}, \textit{skirt}, \textit{pant}). After pre-processing, we obtained 3178 video clips (109K frames), out of which 2579 are used for training and 598 for testing. More details are provided in the supplementary material and Fashion Videos will be made publicly available.

\paragraph{Evaluation metrics}
We evaluate the results via Inception Score (IS)\cite{Salimans2016ImprovedTF}, Fréchet Inception Distance (FID)~\cite{Heusel2017GANsTB}, and Fréchet Video Distance (FVD)~\cite{Unterthiner2018TowardsAG}. Moreover, we modify and use the manipulative precision ($\text{MP}$) metric suggested in~\cite{li2020manigan} to assess the manipulation performance of the models according to the target natural language descriptions. Our version ($\text{MP}_{\text{CLIP}}$) measures the similarity between the manipulated video frames and their corresponding target texts through the cosine similarity in CLIP embedding space~\cite{radford2021learning}. %
Further details on the precise definitions of these metrics are found in our supplementary.

\paragraph{Baselines} As, to the best of our knowledge, the literature lacks a strong~language-driven video manipulation model, we compare 
DiCoMoGAN 
against SISGAN~\cite{dong2017semantic}, TAGAN~\cite{nam2018text} and ManiGAN~\cite{li2020manigan}\footnote{We exclude LightweightGAN~\cite{li2020lightweight} from our analysis as it requires part-of-speech (POS) tagging, and our analysis revealed that existing POS taggers do not give satisfactory results on domain-specific fashion descriptions.}. For video editing, we conduct a frame-by-frame translation. 
\subsection{Evaluations}
\paragraph{Implementation details} For 3D Shapes, we use a 6-dim latent code for the frames in which the first three encode the text relevant static features, the next two the text-irrelevant static features, and the last one the dynamic feature. We set $\beta=32$. For Fashion Videos, we do not have access to the ground truth factors of variation. Thus, we consider a $16$-dim latent code in which the first eight encode the static text relevant features and the next eight the static text-irrelevant ones. The last four are reserved for the dynamic features. We set $\beta=1$.

\begin{figure}[!t]
\centering
\includegraphics[width=\linewidth]{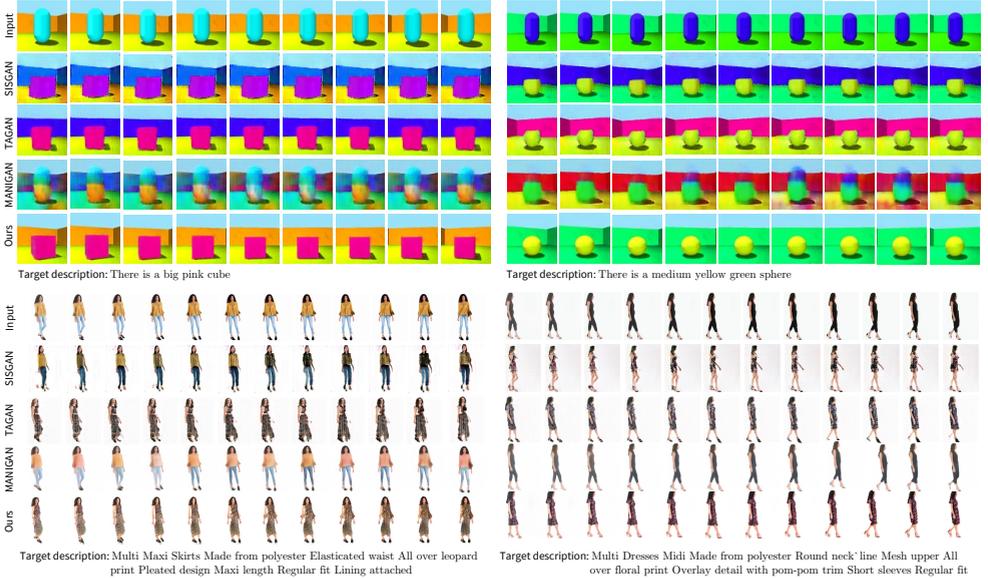}
\caption{\textbf{Qualitative results on 3D Shapes~\cite{3dshapes18} and Fashion Videos datasets}. As compared to SISGAN~\cite{dong2017semantic}, TAGAN~\cite{nam2018text} and ManiGAN~\cite{li2020manigan}, DiCoMoGAN gives sharper images faithful to the target descriptions while preserving inherent features not mentioned in the text, \eg wall and floor colors, identity, hair style, much better than the competing approaches.}
\label{fig:qualitative}
\end{figure}

\begin{table*}[!t]
\centering
\caption{\textbf{Quantitative results.} Our approach outperforms the existing frame-based methods by a large margin in terms of all evaluation measures.}
\setlength{\tabcolsep}{1.5em}
\scriptsize
\begin{tabular}{@{$\;$}c@{$\;\;$}l@{$\;\;$}c@{$\;\;$}c@{$\;\;$}c@{$\;\;$}c}
\toprule
& \textbf{Model} & \bf{IS} ($\uparrow$) & \bf{FID} ($\downarrow$) & \bf{FVD} ($\downarrow$) & \bf{$\text{MP}_{\text{CLIP}}$} ($\uparrow$)\\
\midrule
\parbox[t]{3mm}{\multirow{4}{*}{\rotatebox[origin=c]{90}{\tiny{3D Shapes}}}} & SISGAN~\cite{dong2017semantic} & 2.29 & 138.78& 1185.48& 0.18\\
& TAGAN~\cite{nam2018text} & 2.34 & 88.95 & 974.59 & 0.19\\
& ManiGAN~\cite{li2020manigan} & 2.71 & 26.90 & 753.89 & 0.18\\
& DiCoMoGAN & \textbf{2.76} & \textbf{9.08} & \textbf{69.30} & \textbf{0.26}\\
\bottomrule
\end{tabular}\begin{tabular}{@{$\;$}c@{$\;$}l@{$\;$}c@{$\;$}c@{$\;$}c@{$\;$}c}
\toprule
& \textbf{Model} & \bf{IS} ($\uparrow$) & \bf{FID} ($\downarrow$) & \bf{FVD} ($\downarrow$) & \bf{$\text{MP}_{\text{CLIP}}$} ($\uparrow$)\\
\midrule
\parbox[t]{3mm}{\multirow{4}{*}{\rotatebox[origin=c]{90}{\tiny{Fashion Videos}}}}& SISGAN~\cite{dong2017semantic} & 2.13 & 80.15 & 2274.69 & 0.19\\
& TAGAN~\cite{nam2018text} & 2.24 & 87.60 & 1294.72 & 0.24\\
& ManiGAN~\cite{li2020manigan} & 2.76 & 37.22 & 392.59 & 0.22\\
& DiCoMoGAN & \textbf{2.96} & \textbf{15.34} & \textbf{53.75} & \textbf{0.25}\\
\bottomrule
\end{tabular}
\label{table:results}
\end{table*}

\paragraph{Manipulation results}
In~\cref{table:results}, we provide our quantitative analysis on the 3D Shapes and the Fashion datasets. As compared to the state-of-the-art, our method gives the best results in terms of all of the evaluation metrics. In particular, our method achieves much better FVD values on both datasets. The qualitative results in~\cref{fig:qualitative} indicate that our model can produce high quality results as compared to the existing models. SISGAN and TAGAN fail to preserve the text irrelevant parts like the wall color or the identity of the person. ManiGAN tends to keep the original structure intact and fails to produce the necessary structural changes. Our method, on the other hand, performs more relevant edits on the input video sequences according to the target descriptions, altering only the necessary parts of the frames while keeping what is not mentioned in text unchanged. Please refer to the supplementary material for additional higher resolution results. Our main goal is to analyze disentangling factors in videos for text-based manipulation, and there is definitely room for improvement for visual quality. In the supplementary material, we also show that better results can be obtained when we train a local enhancer network on top of TraNet in a similar vein to pix2pixHD~\cite{wang2018high}.

\begin{wrapfigure}{r}{0.45\textwidth}
\centering
\includegraphics[width=\linewidth]{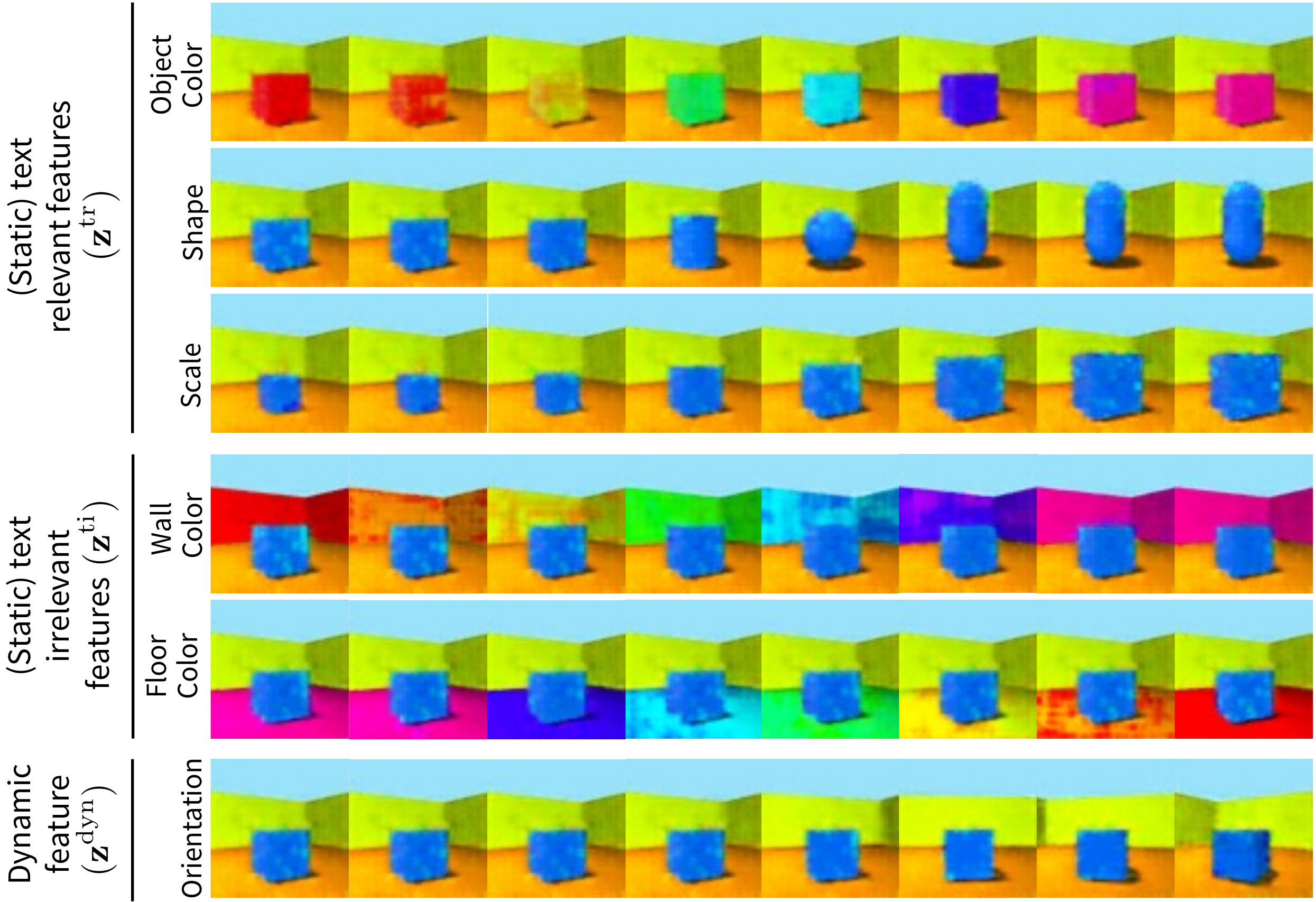}
\caption{\textbf{Latent traversals.}  DiCoMoGAN learns latent variables depicting highly interpretable concepts decomposed into text relevant, text irrelevant static, and dynamic features. Note that wall and floor colors are not mentioned in the descriptions during training.
}
\label{fig:disentanglement}\vspace{-5mm}
\end{wrapfigure}

\noindent\textbf{Measuring disentanglement.}
For disentanglement performance, we carry out experiments on 3D Shapes, where we have ground truth factors of variation, by considering the latent space discovered by RepNet's image encoder. \cref{fig:disentanglement} shows sample traversals in the latent dimensions learned by our method. These traversals clearly depict interpretable 
properties of the images exist in the 3D Shapes dataset. While the last latent dimension steered by the Latent ODE encodes the orientation of the camera (camera movement), all the others encode static features of the scene and the object. In fact, while the first three~static dims. are about text relevant features like object color, shape and size of the object, the last two encode wall and floor colors -- successfully identified without even referred in the provided textual descriptions. We provide further quantitative analysis in the supplementary. 

\noindent\textbf{Ablation study}
In~\cref{tab:ablation}, we show the results of our ablation study in which we examine the contribution of certain components of our method on the performance. Latent ODE within RepNet plays a key role in achieving high-quality manipulation results, which can be attributed to its ability of effectively disentangling motion and content. In the supplementary material, we also analyze the effect of different loss functions and the MFMOD~block.

\begin{table*}[!t]
\renewcommand{\arraystretch}{1.4}
\centering
\caption{\textbf{Ablation study.} Analysis of the components of our DiCoMoGAN model.}
\setlength{\tabcolsep}{5pt}
\scriptsize
\begin{tabular}{c@{$\;$}l@{$\;\;$}c@{$\;\;$}c@{$\;\;$}c@{$\;\;$}c}
\toprule
& \textbf{Model} & \bf{IS} ($\uparrow$) & \bf{FID} ($\downarrow$) & \bf{FVD} ($\downarrow$) & \bf{$\text{MP}_{\text{CLIP}}$} ($\uparrow$)\\
\midrule
\parbox[t]{3mm}{\multirow{3}{*}{\rotatebox[origin=c]{90}{{\tiny{3D Shapes}}}}} & DiCoMoGAN & 2.76 & 9.08 & 69.30 & 0.26\\
& $w/o$ Latent ODE & 2.91 & 12.50 & 99.79 & 0.26\\
& $w/o$ RepNet & 2.82 & 12.32 & 113.89 & 0.25\\
\bottomrule
\end{tabular}\begin{tabular}{c@{$\;$}l@{$\;\;$}c@{$\;\;$}c@{$\;\;$}c@{$\;\;$}c}
\toprule
& \textbf{Model} & \bf{IS} ($\uparrow$) & \bf{FID} ($\downarrow$) & \bf{FVD} ($\downarrow$) & \bf{$\text{MP}_{\text{CLIP}}$} ($\uparrow$)\\
\midrule
\parbox[t]{3mm}{\multirow{3}{*}{\rotatebox[origin=c]{90}{\tiny{Fashion Videos}}}}& DiCoMoGAN & 2.96 & 15.34 & 53.75 & 0.25\\
& $w/o$ Latent ODE & 3.10 & 5.35 & 225.58 & 0.24\\
& $w/o$ RepNet & 3.06 & 13.97 & 498.21 & 0.24\\
\bottomrule
\end{tabular}
\label{tab:ablation}\vspace{-5mm}
\end{table*}

\section{Conclusion and Future Work}
\label{sec:conc}
We presented DiCoMoGAN to tackle the challenging task of manipulation of videos using textual descriptions. As a first step towards solving this problem, we developed a new neural model that incorporates multiple observations to disentangle motion dynamics and visual content to better perform semantically relevant and temporally coherent edits. Our approach gives significantly better results than existing frame-based methods. As such, there are also a number of ways this work could be extended. It is possible to explore more complicated feature aggregation schemes like self-attention~\cite{vaswani2017attention}) to learn permutation invariant representations. Our current model assumes that input video clips include a single object of interest. An exciting future research direction is to incorporate an object-centric approach~\cite{greff2019multi,locatello2020object,emami2021efficient} so that it can support manipulation of multiple objects.

\clearpage

\bibliography{egbib}
\newpage
\section*{Supplementary Material}

%The purpose of this document is to provide extra material to complement our paper ``Disentangling Content and Motion for Text-Based Neural Video Manipulation''. 
We first provide some background material on GANs and their conditional versions, $\beta$-VAE and disentangled representations, Deep Sets, and Neural ODE and Latent ODE, which form the bases for our formulation (Section~\ref{sec:background}). 
 We %
 then give some details about how we collect our Fashion Videos dataset and present the dataset statistics (Section~\ref{sec:fashion}). Next, we explain the architectural choices, training details and hyper-parameter settings of our proposed DiCoMoGAN model along with the evaluation metrics considered in our experimental analysis (Section~\ref{sec:training}). Then, we further demonstrate the disentanglement capabilities of our method through a number of visualizations and comparisons (Section~\ref{sec:visualizations}). We show additional visual comparisons between DiCoMoGAN and the existing frame-based language-guided image editing methods (Section~\ref{sec:additional-qualitative results}). Next, we carry out an ablation study over the loss functions (Section~\ref{sec:additional-quantitative-analysis}). Then, we present sample refinement results obtained by a local enhancement network, giving higher resolution and sharper images (Section~\ref{sec:refinement}). Lastly, we provide some example comparisons against a GAN-inversion based baseline method, which uses CLIP similarity to optimize inverted latent codes of the input frames to align them with the target description (Section~\ref{sec:StyleCLIP}).

 \renewcommand{\thesection}{\Alph{section}}
 \setcounter{section}{0}
\section{Background}\label{sec:background}\vspace{-1mm}

\begin{dfn}[GAN \& Conditional GAN (cGAN)]\textbf{GAN}s~\cite{goodfellow2014generative} implicitly model the distribution of a given training dataset as a 2-player game between a generator and a discriminator network. The generator $G$ maps \mbox{a noise} vector $\z$ sampled from a simple distribution $\Pro_\z$ (\eg a normal distribution)  to a generated sample $G(\z)$, and the discriminator $D$ tries to distinguish real samples $\x \sim \Pro_\x$ from the generated ones $G(\z) \sim \Pro_{G(\z)}$. The parameters of these networks are updated in an alternating manner via the minimax objective given below:
\begin{equation}
    \min _{G} \max _{D}\underset{{\x\sim\Pro_\x}}\E[\log D(\x)]+\underset{\z\sim\Pro_\z}\E[\log (1-D(G(\z)))]
\label{eq:GAN}
\end{equation}

\noindent As a special instance of this adversarial model, \textbf{cGAN}s~\cite{mirza2014conditional} consider  where each training sample is associated with some extra information (condition) (\eg class label~\cite{odena2017conditional,nguyen2017plug,brock2018large}, text ~\cite{reed2016generative,zhang2017txt2im} or an image or a layout~\cite{isola2017image,yang2018diversity,wang2018high,park2019semantic}), and the task is to model the conditional distribution $\Pro_{\x\mid \y}$ from a training set of sample pairs $\{(\x_i,\y_i)\}_{i=1}^N\subseteq\gX\times\gY$ drawn from the joint distribution $\Pro_{\x\y}$ of observations $\x$ and the corresponding conditions $\y$. Differently from GANs, in cGANs, the condition information is also fed to both the generator and the discriminator, resulting in the objective:
\begin{equation}
    \min _{G} \max _{D}\underset{(\x,\y)\sim\Pro_{\x\y}}\E[\log D(\x,\y)]+\underset{\z\sim\Pro_\z,\y\sim\Pro_\y}\E[\log (1-D(G(\z,\y),\y))]
\label{eq:cGAN}
\end{equation}
\end{dfn}

\begin{dfn}[$\beta$-VAE \& Disentangled Representations] 
To learn the hidden low-dimensional latent structure of high-dimensional observations, the \textbf{$\boldsymbol{\beta}$-VAE} model~\cite{Higgins2017betaVAELB} extends the VAE framework~\cite{kingma2014auto,rezende2014stochastic} by penalizing the standard $\KL$ divergence term between the variational posterior $q_{\phi}$ and the prior $p(\z)$ (typically fixed) with a larger weight ($\beta \ge 1$) (the vanilla VAE has $\beta=1$), as: 
\begin{equation}
    \E_{q_{\phi}(\z\vert\x)}\left[\E_{p(\x)}\left[\log p_{\theta}(\x\vert\z)\right]-\beta \KL\left(q_{\phi}(\z\vert\x) \| p(\z)\right)\right]
\label{eq:betaVAE}
\end{equation}
where $q_\phi(\z\vert\x)$ is the encoder that maps a sample $\x$ to a latent code $\z$ and $p_\theta(\x\vert\z)$ is the probabilistic decoder that can generate observations from seen/unseen latent codes, which are both parameterized by neural networks with learnable parameters $\phi$ and $\theta$, respectively. 
Such modification of the VAE objective (ELBO) is shown effective for learning \textbf{disentangled representations}~\cite{bengio2013representation} as the $\beta$ value helps to achieve less correlation between the latent dimensions, resulting semantically more meaningful latent codes~\cite{Higgins2017betaVAELB,burgess2018understanding,mathieu2019disentangling}. 
\end{dfn}

\begin{dfn}[Deep Sets] 
Let $S_n$ be the set of all permutations of indices $\{1,...,n\}$. A function \mbox{$f:\gX^n \rightarrow \gY^n$} is \textbf{permutation invariant} iff for any permutation $\pi \in S_n$, $f(\pi(X)) = f(X)$, where $X=\{\x_i\}_{i=1}^n\in \gX^n$ denotes a set consisting of $n$ items. \textbf{Deep Sets}\cite{Zaheer2017DeepS} proves that such a function $f$  defined over sets can be implemented with learnable neural network modules by decomposing it to the form $f(X)=\rho(\sum_{\x\in X} \psi(\x))$, where $\psi$ is an encoder network that extracts a feature embedding of each item, $\rho$ is a decoder network that acts on aggregated features computed by the summation over the set elements. Note that other pooling functions such as $\max(\cdot)$ operation can also be employed to enforce permutation invariance~\cite{ravanbakhsh2016deep,Zaheer2017DeepS,qi2017pointnet}.
\end{dfn}

\begin{dfn}[Neural ODE \& Latent ODE]\label{dfn:latentODE}
We construct an ordinary differential equation (\textbf{ODE}) by attaching a time dependent vector $f(\z, t)\in\R^d$ to every point $\z\in\Man\subset\R^d$ resulting in a vector field s.t. $\frac{d{z(t)}}{dt} = f(z(t),t)$, given the initial state $\z_0$. This ODE can be integrated for time $T$ modeling the~flow:%
\begin{equation}
\label{eq:ode}
    \z_T = \phi_T(\z_0) = \z_0 + \int_{0}^T f_{\theta}(\z_t, t) \, dt,
\end{equation}
where $\z_t\triangleq z(t)$ and the field $f:\R^d\mapsto\R^d$ is parameterized by $\ODEpars=\{\theta_i\}_i$. By the Picard–Lindelöf theorem~\cite{coddington1955theory}, if $f$ is continuously differentiable then the initial value problem in~\cref{eq:ode} has a unique solution. 
Instead of handcrafting, \textbf{Neural ODEs}~\cite{chen2018neural} seek a function $f$ that suits a given objective by modeling $f$ as a neural network.
This brings several advantages such as having theoretically infinite number of layers, memory efficiency thanks to the use of the adjoint sensitivity~\cite{pontryagin2018mathematical}, exact invertibility, continuity and smoothness.
Numerous forms of Neural ODEs model $f(\cdot)$ to be 'autonomous' \ie a time independent function $f(\z_t)\equiv f(\z_t, t)$~\cite{chen2018neural,dupont2019augmented,rubanova2019latent}, whose output fully characterizes the trajectory.
We refer to a Neural ODE operating in a latent space as a \textbf{Latent ODE}.\vspace{-3mm}
\end{dfn}

\section{Fashion Videos Dataset}
\label{sec:fashion}
\begin{figure}[!h]
    \centering
    \includegraphics[width=\linewidth]{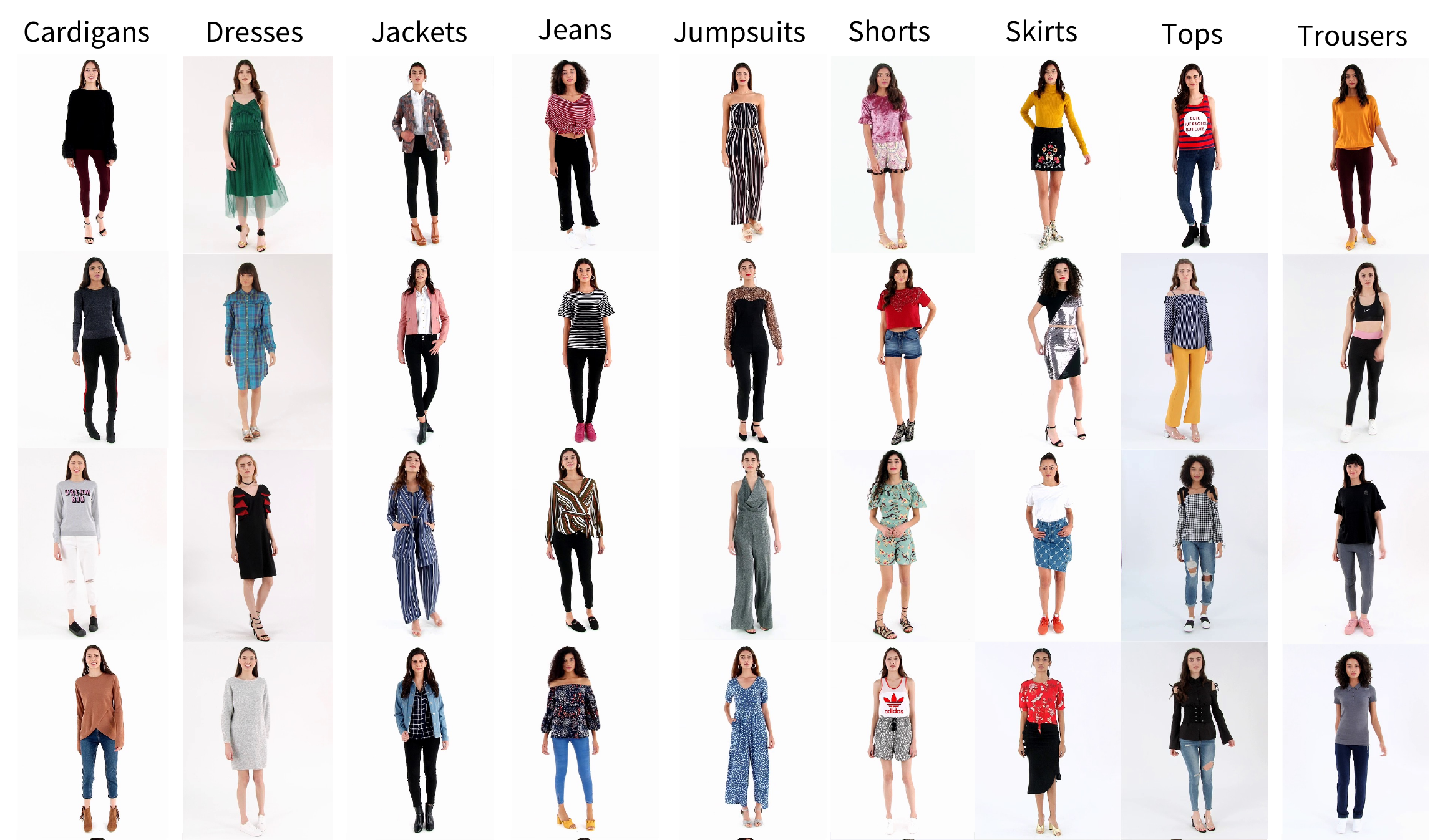}
    \caption{\textbf{Fashion Videos dataset.} Sample frontal frames showing different garment types in our Fashion Videos dataset. Please refer to the accompanying project web page for the full video examples.}
    \label{fig:fashion-samples}
\end{figure}

\paragraph{Data Collection} We collected our dataset from raw videos present in the website of an online clothing retailer by searching products in the \textit{cardigans, dresses, jackets, jeans, jumpsuits, shorts, skirts, tops} and \textit{trousers} categories. Fig.~\ref{fig:fashion-samples} illustrates these nine garment types and the variety within each category with some samples. Table~\ref{tab:fashion} presents dataset statistics per each category. After extracting the video frames and resizing them to 533$\times$400 resolution, we manually removed the frames containing zooms to the item, which generally appear at the end of the video clips. As stated in the main text, this resulted in 3178 video clips (approximately 109K distinct frames), which we split into 2579 for training and 598 for testing.

\begin{table}[!h]
    \caption{\textbf{Dataset statistics.} The number of video sequences per each garment category in the Fashion Videos dataset.}
    \centering
    \begin{tabular}{l@{$\;\;$}r@{$\qquad$}l@{$\;\;$}r@{$\qquad$}l@{$\;\;$}r}
    \toprule
     \textbf{Category} & \textbf{Count} & \textbf{Category} & \textbf{Count} & \textbf{Category} & \textbf{Count} \\
    \midrule
         \textit{trousers} & 263 & \textit{shorts} & 26 & \textit{jackets} & 122\\ 
         \textit{tops} & 1495 & \textit{jumpsuits} & 122 & dresses & 853\\
         \textit{skirts} & 84 & \textit{jeans} & 78 & \textit{cardigans} & 138\\ 
    \bottomrule
    \end{tabular}
    \label{tab:fashion}
\end{table}

\paragraph{Garment descriptions} We obtained textual descriptions of the clothes from the headers of the html files by extracting the hierarchy and info sections of the items. These product descriptions give details about its color, material properties and design details. We only manually removed the repetitive or ill-suited words from the descriptions.  In addition to that, we shuffled the ordering of the words whenever it is linguistically possible to augment textual data, which gave us roughly six different descriptions for each video clip.

\section{Training, Architecture, and Evaluation Details}
\label{sec:training}

\paragraph{Architecture Details}
In Table~\ref{tab:details}, we give the architecture details of each component of our DiCoMoGAN video manipulation framework. The table follow a naming convention similar to the one used in~\cite{zhu2017unpaired,wang2018high}. In particular,
we use $\mathtt{CN}$,$\mathtt{C}$ and $\mathtt{L}$ for Convolution-InstanceNorm-ReLU, Convolution-ReLU and Linear layers, respectively. For example, $\mathtt{CN_{3}256S_2}$ denotes a Convolution-InstanceNorm-ReLU layer with $256$ filters of kernel size $3\times3$ kernel and stride $2$. If different stride is used for each spatial dimension, we denote it with vector like $\mathtt{S_{2,1}}$.  $\mathtt{f_{i}}$ represents $i$-th internal feature tensors. $\mathtt{MF512}$ denotes proposed multi-feature modulation block (MFMOD) with filter size $512$. In TraNet, each of these MFMOD blocks performs feature modulation by considering both the text  $\w^{\mathrm{desc}}\in\sR^{512}$ and the content $\w^{\mathrm{cont}}\in\sR^{256}$ features denoted with $\mathtt{A[}  \w^{\mathrm{desc}},  \w^{\mathrm{cont}} \mathtt{]}$. $\w^{\mathrm{cont}}$ is computed from text irrelevant and dynamic latent codes through a mapping network which is 4-layer MLP network.   $\mathtt{D_{3}128S_{0.5}}$ represents a Deconvolution-InstanceNorm-ReLU layer with $128$ filters of kernel size $3\times3$ and stride $0.5$. At the last deconvolution layer $\mathtt{D_{7}3S_1}$, we do not use InstanceNorm and replace ReLU activations with $\mathrm{tanh}$.

\paragraphnoper{The representation network (RepNet)} uses an $\beta$-VAE like encoder-decoder architecture  with additional components, a Neural ODE $f_\textrm{ODE}$, a $\mathrm{GRU}$ module and an aggregation function $\mathrm{\max}$. First, the encoder network maps input images $\mathbf{x}^0,...,\mathbf{x}^k$ to hidden features $\mathbf{h}^0,...,\mathbf{h}^k$. Then $\mathrm{\max}$ function aggregate those  hidden features to construct static latent codes $\mathbf{z}^{ST}$. On another path, $\mathrm{GRU}$ module takes same hidden features and time gaps between time stamps of input frames in reverse order to produce $\mathbf{z}^{\mathrm{dyn}}_{t_0}$ which is then passed to $f_\textrm{ODE}$ to predict dynamic latent codes $\mathbf{z}^{\mathrm{dyn}}_{t}$ on given time stamps. In our experiments, we use a feed-forward neural network~\cite{chen2018neuralode,chen2021eventfn} as ODE function $f_{\mathrm{ODE}}$, which consists of $3$ fully connected layer with ELU activation function. Our ODE Solver is \texttt{\small torchdiffeq} Python package with fifth order \texttt{\small dopri5} solver and adaptive step for generative modeling. Relative and absolute tolerances are set as 1E-7 and  1E-9, respectively. Moreover, an auxiliary text encoder takes in text features $\w^{\mathrm{desc}}$ as input and projects it to text relevant latent code $\textbf{z}^{\mathrm{desc}}$. Finally, concatenated static and dynamic latent codes are given to the decoder to reconstruct input images. Expected text relevant parts of static latent code are changed with $\textbf{z}^{\mathrm{desc}}$ on the second pass from decoder for a second reconstruction. Note that the disentanglement loss $\mathcal{L}^\mathrm{dyn}_\mathrm{KL}$ (Eqn. 9) is defined over $\mathbf{z}_{t_0}^{\mathrm{dyn}}$ and $\mathbf{x}_{t_0}$. This comes from the assumption of Latent ODEs~\cite{rubanova2019latent} that the continuous trajectories are deterministic functions of their initial states. Similarly, in $\mathcal{L}^\mathrm{ST}_\mathrm{KL}$, we average over the observations so that the losses involving static and dynamic features are balanced with respect to each other.

\paragraphnoper{The translation network (TraNet)} is responsible for manipulating an input video frame with respect to the given target description. It contains a content encoder that produces $512$ dimensional spatial features for an input video frame $\mathbf{x}^{t_i}$. These features are then passed to a series of the proposed multi-feature modulation (MFMOD) blocks. As explained in detail in Sec. 4.1, each MFMOD block modulates its input features by the text $\w^{\mathrm{desc}}$ and the content $\w^{\mathrm{cont}}$ codes derived from RepNet. Output features of the last MFMOD block is fed to the content decoder to generate the manipulated frame.

\paragraphnoper{The discriminator network} takes the real/generated image $\mathbf{x}$ / $\mathbf{y}$ as input along with the matched / unmatched / relevant target combinations of text  $\w^{\mathrm{desc}}$ and  content  $\w^{\mathrm{cont}}$ features. A $\mathtt{MF512}-\mathtt{A[}  \w^{\mathrm{desc}},  \w^{\mathrm{cont}} \mathtt{]}$  block modulates image features features before classification layer. We use leaky ReLU ($\mathtt{LReLU}$) with slope $0.2$ for our discriminator network. We do not use InstanceNorm at the input layers. Moreover, we employ $3$ discriminator models at $3$ different spatial scales with $1, 0.5$ and $0.25$ as the scaling factors during training. We trained our main network for roughly $200$ epochs on a NVIDIA Quadro RTX 8000 GPU for $7$ days for fashion dataset. 

\begin{figure}[!h]
    \centering
    \includegraphics[width=\linewidth]{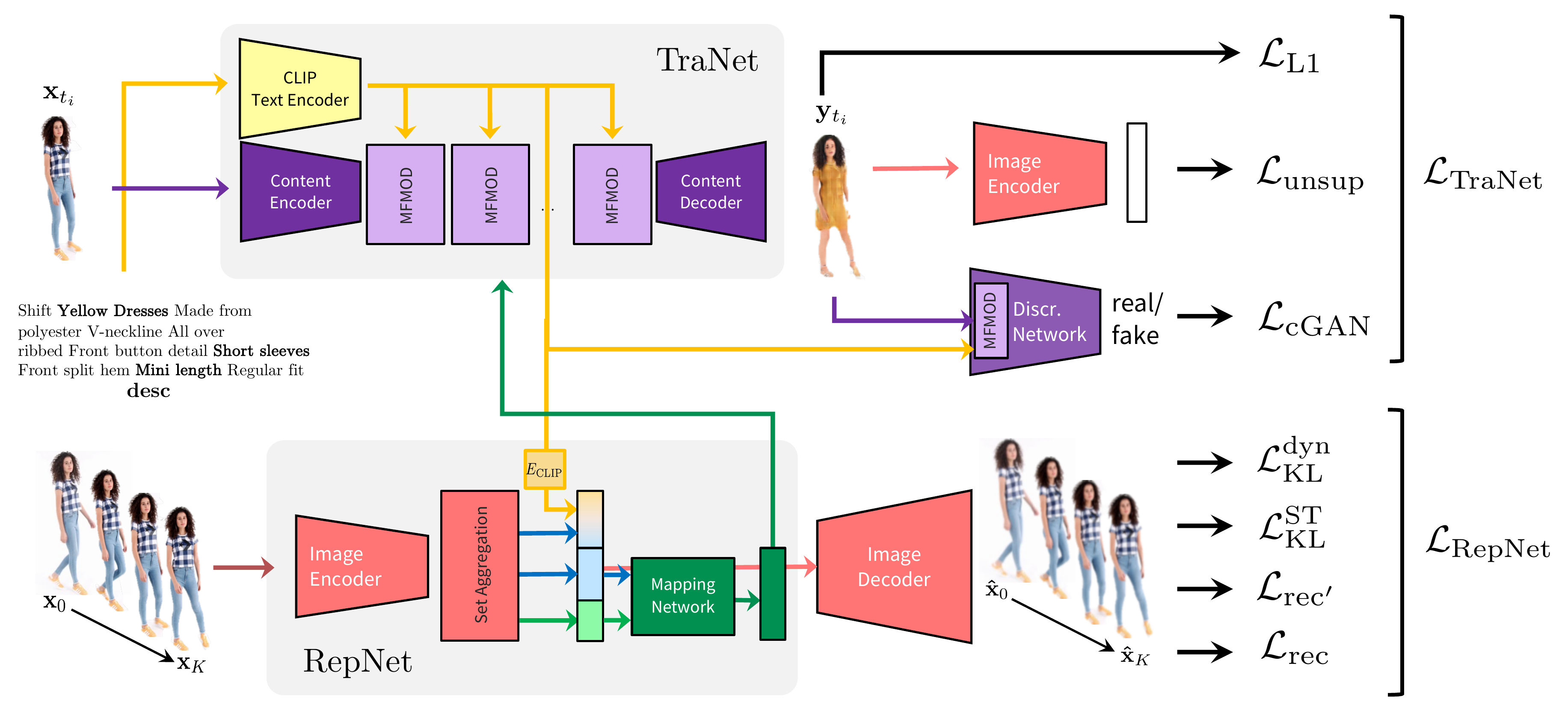}
    \caption{\textbf{A detailed look at training of our DiCoMoGAN framework}.}
    \label{fig:training}
\end{figure}

Fig.~\ref{fig:training} shows an overall flow of training of our DiCoMoGAN frameworks, together with all of its components and the corresponding loss functions. 

\begin{table}[h!]
\centering
\caption{\textbf{Architecture details of DiCoMoGAN}. Our proposed DiCoMoGAN model consists of a translation network (TraNet), a representation network (RepNet), a discriminator network, and a mapping network. Please refer to the text for the detailed explanations of the abbreviations used in the specifications.}  
\resizebox{\linewidth}{!}{
\begin{tabular}{ll}
\hline%
\multicolumn{2}{c}{\textbf{Translation Network (TraNet)}} \\
\hline
\textbf{Input} & \textbf{Specification} \\
\hline
$\mathbf{x}$ & $\mathtt{CN_{7}64S_1-CN_{3}128S_2-CN_{3}256S_2-CN_{3}512S_2\rightarrow \mathbf{f}_1}$ \\
$\mathbf{f}_1, \w^{\mathrm{cont}} , \w^{\mathrm{desc}} $ & $\mathtt{MF512}-$$\mathtt{A[}  \w^{\mathrm{desc}},  \w^{\mathrm{cont}} \mathtt{]}-$$\mathtt{MF512}-$$\mathtt{A[}  \w^{\mathrm{desc}},  \w^{\mathrm{cont}} \mathtt{]}-\mathtt{MF512}-$$\mathtt{A[}  \w^{\mathrm{desc}},  \w^{\mathrm{cont}} \mathtt{]}-$\\
& $\mathtt{MF512}-$$\mathtt{A[}  \w^{\mathrm{desc}},  \w^{\mathrm{cont}} \mathtt{]}-\mathtt{MF512}-$$\mathtt{A[}  \w^{\mathrm{desc}},  \w^{\mathrm{cont}} \mathtt{]} \rightarrow \mathbf{f}_2$ \\
$\mathbf{f}_2$&$\mathtt{D_{3}256S_{0.5}-D_{3}128S_{0.5}-D_{3}64S_{0.5}\rightarrow \mathbf{f}_{3}}$\\
$\mathbf{f}_{3}$&$\mathtt{D_{7}3S_1}\rightarrow \mathbf{y}$ \\
\hline
\multicolumn{2}{c}{\textbf{Mapping Network}} \\
\hline
\textbf{Input} & \textbf{Specification} \\
\hline
$\mathbf{z^{ti}}, \mathbf{z^{dyn}_{t}}$&$\mathtt{L_{256}-LReLU-L_{256}-LReLU-L_{256}-LReLU-L_{256}\rightarrow \textbf{w}^{\mathrm{cont}}}$\\
\hline
\multicolumn{2}{c}{\textbf{Representation Network (RepNet)}} \\
\hline
\textbf{Input} & \textbf{Specification} \\
\hline
$\mathbf{x}^0,...,\mathbf{x}^k$& $\mathtt{C_{4}32S_2-C_{4}32S_2-C_{4}64S_2-C_{4}128S_2-C_{4}128S_2 - L_{256}-ReLU \rightarrow}$ $\mathbf{h}^0,...,\mathbf{h}^k$\\
$\mathbf{h}^0,...,\mathbf{h}^k$&$\mathtt{MAX - L_{12} \rightarrow \mathbf{z}^{ST}}$\\
$\mathbf{h}^0,...,\mathbf{h}^k, \Delta t_0,...,\Delta t_k $&$\mathtt{GRU_{256} - L_{4} \rightarrow \mathbf{z}_{t_0}^{\mathrm{dyn}}}$\\
$\mathbf{z}_{t_0}^{\mathrm{dyn}}, t_0,...,t_k$&$\mathtt{\mathrm{ODESolve}(\fnode, \z_{t_0}^{\mathrm{dyn}}, (t_0, t_1, ..., t_{K})) \rightarrow \mathbf{z}^{\mathrm{dyn}}_{t_0},...,\mathbf{z}^{\mathrm{dyn}}_{t_k}}$\\
$\textbf{w}^{\mathrm{desc}}$ &$\mathtt{L_{256}-LReLU-L_{128}-LReLU-L_{64}-LReLU-L_{8}\rightarrow \textbf{z}^{\mathrm{desc}}}$\\
$\mathbf{z}^{ST}, \mathbf{z}^{\mathrm{dyn}},\textbf{z}^{\mathrm{desc}}$&$\mathtt{L_{256} - ReLU - L_{128 \times 4 \times 3} -ReLU - D_{4}128S_{0.5}-D_{4}64S_{0.5}-D_{4}32S_{0.5}-D_{4}32S_{0.5}-}$\\
&$\mathtt{D_{4}3S_{0.5}} \rightarrow \mathtt{\mathbf{\hat{x}}^0,...,\mathbf{\hat{x}}^k}$\\
\hline
\multicolumn{2}{c}{\textbf{Discriminator Network}}\\
\hline
\textbf{Input} & \textbf{Specification} \\
\hline
$\mathbf{x} | \mathbf{y}$ & $\mathtt{CN_{4}64S_2-CN_{4}128S_2-CN_{4}256S_2-CN_{4}512S_2 \rightarrow \mathbf{f}_1}$\\
$\mathbf{f}_1, \w^{\mathrm{cont}}, \w^{\mathrm{desc}}$ & $\mathtt{MF512}-$ $\mathtt{A[}  \w^{\mathrm{desc}},  \w^{\mathrm{cont}}\mathtt{]}\mathtt{- C_{4}1S_1} \rightarrow \{0,1\}$\\
\hline
\end{tabular}
}
\label{tab:details}
\end{table}

\paragraph{Training Details}
In our experiments, we used Adam optimizer with $\beta_1=0.5$ and $\beta_2=0.999$ and the learning rate of $0.0002$ for TraNet and the discriminator network. For the mapping network, the learning rate is decreased by two orders of magnitude ($0.0002/100 $) as in StyleGAN~\cite{karras2019style} to stabilize the training process. For RepNet, we used Adam optimizer with $\beta_1=0.9$ and $\beta_2=0.999$ and the learning rate of $0.001$. We defined separate optimizers for RepNet, TraNet and the discriminator network, and due to the GPU memory limitation, we set the batch size to 32 and 8 for the 3D Shapes and the Fashion datasets, respectively. We trained our models including ours for $100$ and $200$ epochs on the 3D Shapes and the Fashion Videos datasets, respectively. We set $\beta=32$ on 3D Shapes and $\beta=1$ on Fashion for $\beta$-VAE and we picked the first frame and $3$ random video frames in increasing time order to obtain multiple observations for both datasets. Lastly, SISGAN~\cite{dong2017semantic}, TAGAN~\cite{nam2018text} and ManiGAN~\cite{li2020manigan} are trained for same number of epochs with ours by keeping their original settings as much as possible. 

\paragraph{Evaluation Metrics}
In our quantitative analysis, we evaluate our model in three different aspects, namely photorealism, relevance to the target description, and disentanglement performance. In particular, to assess the photorealism of the edited videos, we employ Inception Score (IS) \cite{Salimans2016ImprovedTF}, Fréchet Inception Distance (FID)~\cite{Heusel2017GANsTB}, and Fréchet Video Distance (FVD)~\cite{Unterthiner2018TowardsAG}. 

\paragraphnoper{Inception Score (IS)}~\cite{Salimans2016ImprovedTF} is built on the assumption that a good generative model should generate diverse samples such that their semantic classes can be confidently predicted by a visual classifier. It measures this by estimating the KL-divergence between the conditional class distribution $p(y|\x)$ and the marginal class distribution $p(y)$:
\begin{equation}
    \mathrm{IS}(G) = \mathrm{exp}\left( \mathbb{E}_{\x\sim p_G}\left[D_{\mathrm{KL}}(p(y|\x)||p(y))\right]\right)
\end{equation}
\noindent with $\x\sim p_G$ denoting a synthetic image sampled from the generative model $G$, $D_{KL}(p||q)$ representing the KL-divergence between the distributions $p$ and $q$. As its name implies, the classifier used to obtain the class predictions is an Inception Network~\cite{szegedy2016inception} pre-trained on ImageNet. A high IS represents a good visual quality. 

\paragraphnoper{Fréchet Inception Distance (FID)}~\cite{Heusel2017GANsTB} provides an alternative to IS in evaluating the photorealism of the generated synthetic images. It estimates the statistical distance between the distribution of real and the distribution of synthetic images by casting them as multivariate Gaussians:
\begin{equation}
    \mathrm{FID}(G) = |\mu_{R}-\mu_{G}|^2+\text{Tr}\left(\Sigma_R+\Sigma_G-2\left(\Sigma_R\Sigma_G\right)^{1/2}\right)
\end{equation}

\noindent Here, $\mu_{R}$ and $\mu_{G}$ represent the means of the real world data distribution and the distribution defined the generative model, respectively. Similarly, $\Sigma_R$ and $\Sigma_G$ denote the co-variance matrices, respectively. To obtain the feature representations of real and synthetic images, again a pre-trained Inception Network~\cite{szegedy2016inception} is used, \emph{i.e.} each image is represented with the activations of one of the hidden layers of a pre-trained Inception Network. Lower FID value means that the distributions of generated images and real images are similar to each other. 

\paragraphnoper{Fréchet Video Distance (FVD)}~\cite{Unterthiner2018TowardsAG} is just an extension of FID for videos. In particular, FVD replaces the Inception Network with the Inflated 3D Convnet (I3D) model~\cite{carreira2017i3d}, which employs 3D convolutions to process videos and which is pre-trained on the Kinetics action recognition dataset including human-centered videos from YouTube. By this way, it takes into account the temporal coherence of the visual content across a given video sequence as well while obtaining its feature representation. Hence, it is better suited for evaluating the quality of the generated videos. Like FID, higher the FVD, the better the visual quality. 

\paragraphnoper{CLIP-based Manipulative Precision ($\textbf{MP}_\textbf{CLIP}$)} measures the manipulation performance of a language-based image editing method. It is a modified version of the manipulative precision ($\text{MP}$) metric~\cite{li2020manigan}, which is defined as follows:
\begin{equation}
    \text{MP}_{\text{CLIP}}(\y,\x,\w) = \left(1-|\y-\x|_1\right)\times \text{sim}_{\text{CLIP}}(\y,\w)
\end{equation}
\noindent with $\x$ and $\w$ representing an input image and a target description, respectively, and $\y$ denoting the editing outcome. An ideal manipulation model should make only the minimal changes over the input image as suggested by the target description. In other words, it should only make the necessary modifications on the visual content by keeping the text irrelevant parts intact. In that respect, $\text{MP}_{\text{CLIP}}$ combines the relevance of the edited videos with the given target descriptions with a term that measures the pixel difference. In particular, the relevance is computed by the cosine similarity between the manipulated image and the target description in the CLIP embedding space~\cite{radford2021learning}. We estimate this score for every frame in a given video sequence, and then report their average. A high $\text{MP}_{\text{CLIP}}$ score indicates a good manipulation performance. 

\paragraphnoper{Mutual Information Gap (MIG)}~\cite{chen2018isolating} is an information theoretic metric proposed to assess the disentanglement performance of a latent encoder model by considering the empirical mutual information between a latent variable $z_j$ and a ground truth factor $v_k$ as given by:
\begin{equation}
    I_{x_n}(z_j,v_k)=\mathbb{E}_{q(z_j,v_k)}\left[\text{log}\sum_{x_n}q(z_j|x_n)p(x_n|v_k) \right]+H(z_j)
\end{equation}
with $H(x)$ representing the entropy function and $q(z_j,v_k)$ denoting the joint distribution:  
\begin{equation}
    q(z_j,v_k)=\sum_{n=1}^N p(v_k)p(x_n|v_k)q(z_j|x_n)
\end{equation}
where $p(v_k)$ models the underlying factors, $p(n|v_k)$ represents the generating process for the samples, and $q(z_j|n)$ denotes the inference distribution. Then, $\text{MIG}$ score is estimated by first computing the empirical mutual information for every pair of $z_j$ and $v_k$, and then by measuring the difference between the top two latent variables with highest mutual information:
\begin{equation}
    \text{MIG}(\mathbf{v},\z)=\frac{1}{K}\sum_{k=1}^K \frac{1}{H(v_k)}\left( I_n(z_{j^(k)},v_k)-\text{max}_{j\neq j^{(k)}}I_n(z_j,v_k)\right)
\end{equation}
with $j^{(k)}=\text{argmax}_j I_n(z_j,v_k)$, $K$ denoting the number of known factors. $\text{MIG}$ is bounded by 0 and 1, and a higher MIG score represents a better disentanglement of the latent codes.

\paragraphnoper{Axis Alignment Metric (AAM)}~\cite{aam} is another quantitative measure of disentanglement that specifically takes into account the axis alignment of the latent variables. Formally, it is defined as follows:
\begin{equation}
    \text{AAM}(\mathbf{v},\z)=\frac{1}{K}\sum_{k=1}^K \frac{\max\left(\text{max}_jI_{x_n}(z_j,v_k)-\sum_{j'=1}^{d-1}I_{x_n}(z_j,v_k)_{j'},0\right)}{\text{max}_jI_{x_n}(z_j,v_k)} 
\end{equation}
where $d$ represents the $d$-th order statistics. While MIG measures on how much information of $\mathbf{v}$ is encoded by $\z$, AAM examines whether each groundtruth factor of variation $v_k$ is being encoded in a single latent variable $z_j$ or not. Higher AAM scores indicate better disentanglement of the latent variables.

\section{Disentanglement Analysis}
\label{sec:visualizations}
\paragraph{Quantitative Evaluation} In~\cref{tab:disentanglement}, we compare our method against two different $\beta$-VAE models, which are trained with and without a Latent ODE, and a modified version of our model that lacks the Latent ODE component. The quantitative results suggest that performing a joint training over the reconstruction and translation losses lead to better AAM scores. Morever, introducing the Latent ODE improves the disentanglement by explicitly separating the latent representation into static and dynamic parts. Overall, our DiCoMoGAN model gives the highest disentanglement scores.
\renewcommand{\arraystretch}{1.15}
\begin{table}[!t]
\caption{\textbf{Disentanglement evaluation.} Quantitative disentanglement comparisons on 3D Shapes~\cite{3dshapes18}. Higher values are better for MIG and AAM. MIG scores are multiplied by 10$^3$. All models use $\beta=32$.}
\centering
\setlength{\tabcolsep}{15pt}
\footnotesize
\begin{tabular}{l@{$\quad$}c@{$\quad$}c}
\midrule
\textbf{Model} & \textbf{MIG} ($\uparrow$) & \textbf{AAM} ($\uparrow$)\\
\midrule
$\beta$-VAE & 0.82 & 0.46\\
$\beta$-VAE + Latent ODE & 0.90 & 0.64\\
\midrule
DiCoMoGAN & \textbf{3.56} & \textbf{0.96}\\
DiCoMoGAN $-$ Latent ODE & 0.69 & 0.71\\
\midrule
\end{tabular}
\label{tab:disentanglement}
\vspace{-2mm}
\end{table}

\paragraph{Traversals on 3D Shapes}
As reported in~\cref{tab:disentanglement}, our DiCoMoGAN beats $\beta$-VAE and $\beta$-VAE-ODE baselines in terms of disentanglement performance based on both MIG~\cite{chen2018isolating} and AAM~\cite{aam} metrics. This is expected as the RepNet component of DiCoMoGAN utilizes additional weak supervision through the adversarial training of TraNet. In Fig.~\ref{fig:disentanglement}, we qualitatively show their differences in the disentanglement using visualizations of the latent traversals on 3D Shapes~\cite{3dshapes18}. As can be seen, the latent factors identified by DiCoMoGAN are independent than each other and encode distinct features, which is not the case for $\beta$-VAE and $\beta$-VAE-ODE models. For instance, in $\beta$-VAE, wall color also changes when traversing over the second and the last dimensions, which empirically correspond to shape and orientation, respectively. Compared to $\beta$-VAE, disentangling motion and content via a latent ODE ($\beta$-VAE-ODE) provides better disentanglement, but still some dimensions are mixed with shape like object color and wall color. 

\begin{figure}[!h]
\centering
\resizebox{\linewidth}{!}{
\begin{tabular}{c@{$\;$}c@{$\;$}c}
\includegraphics[height=0.25\linewidth]{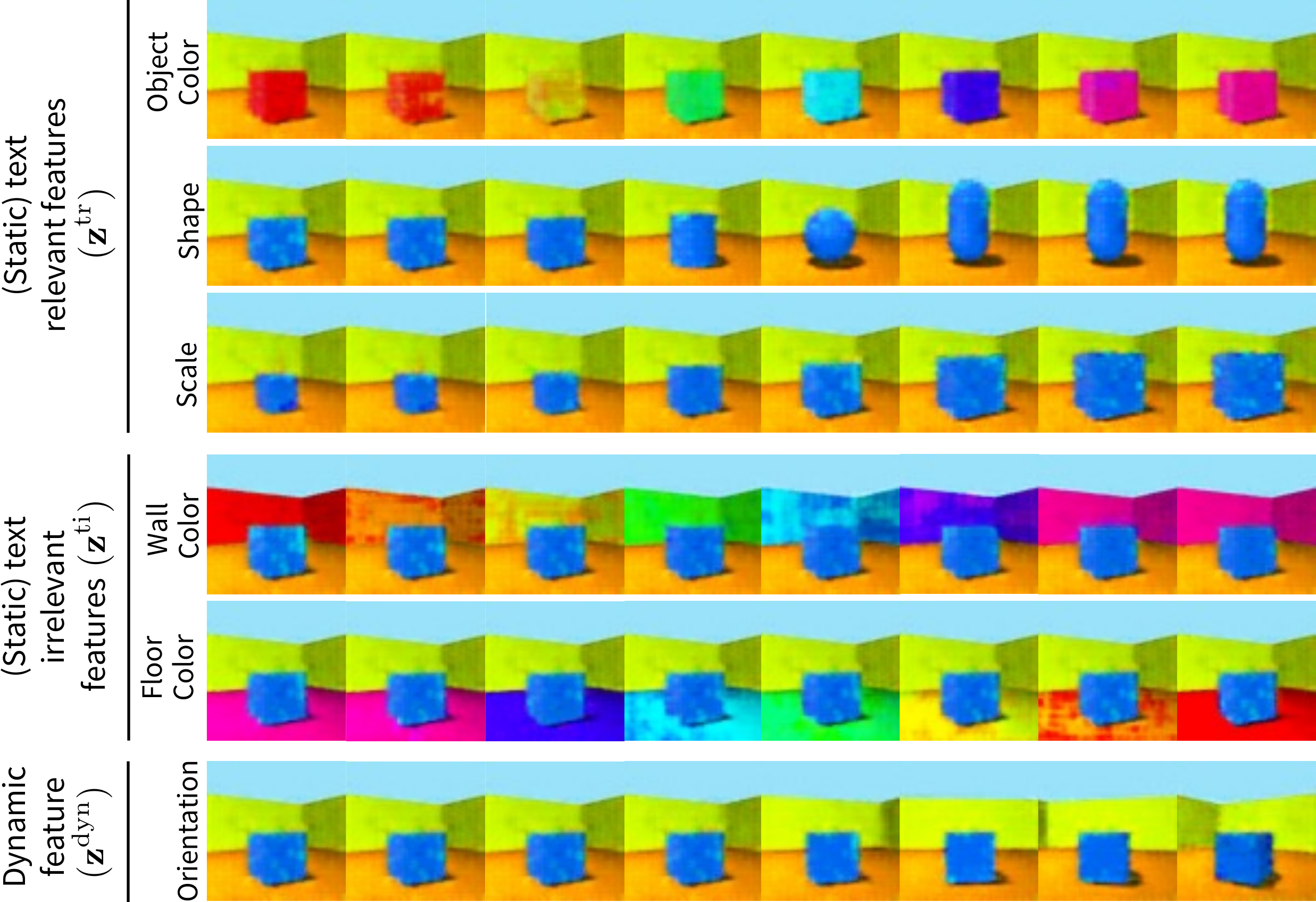} &
\includegraphics[height=0.25\linewidth]{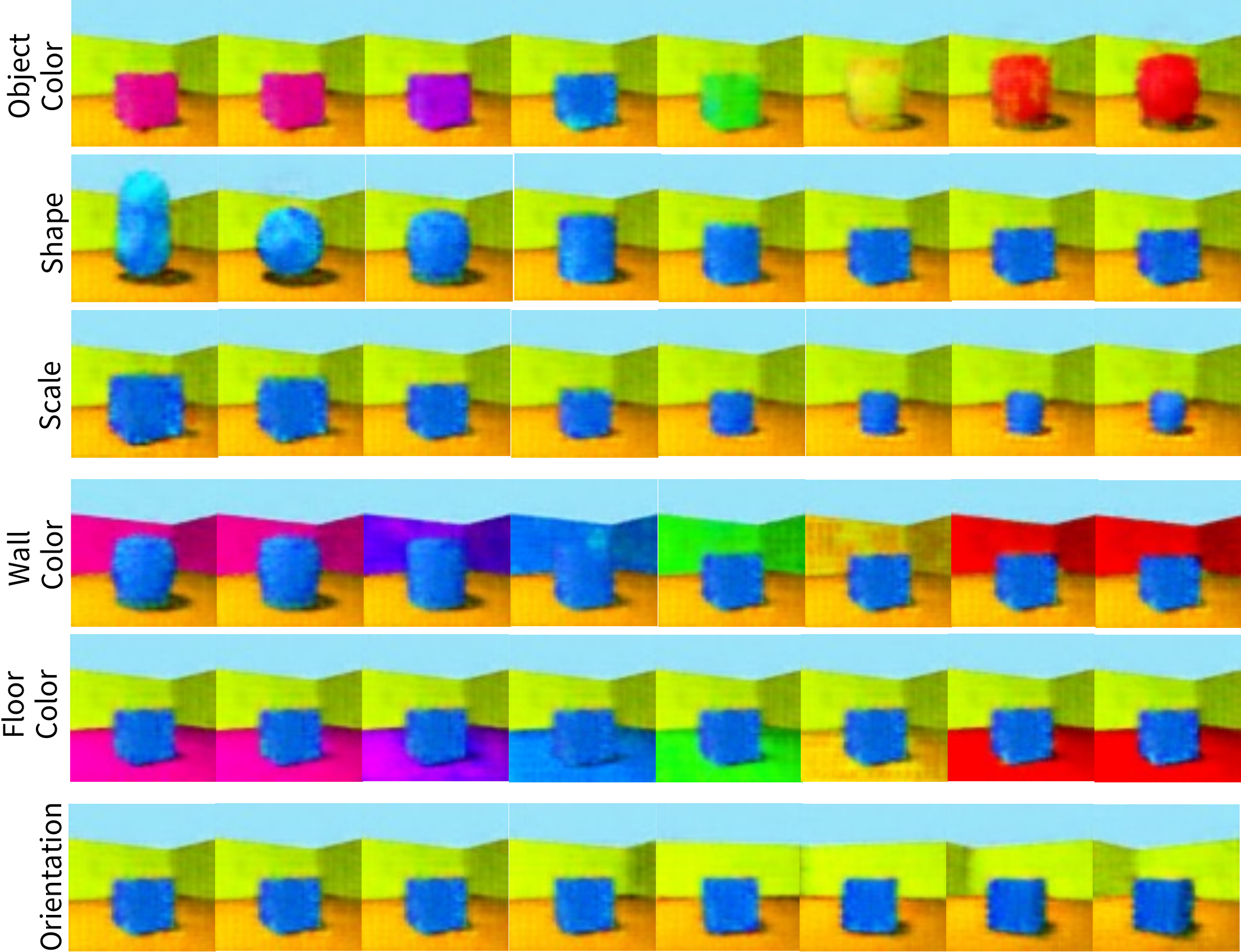} & \includegraphics[height=0.25\linewidth]{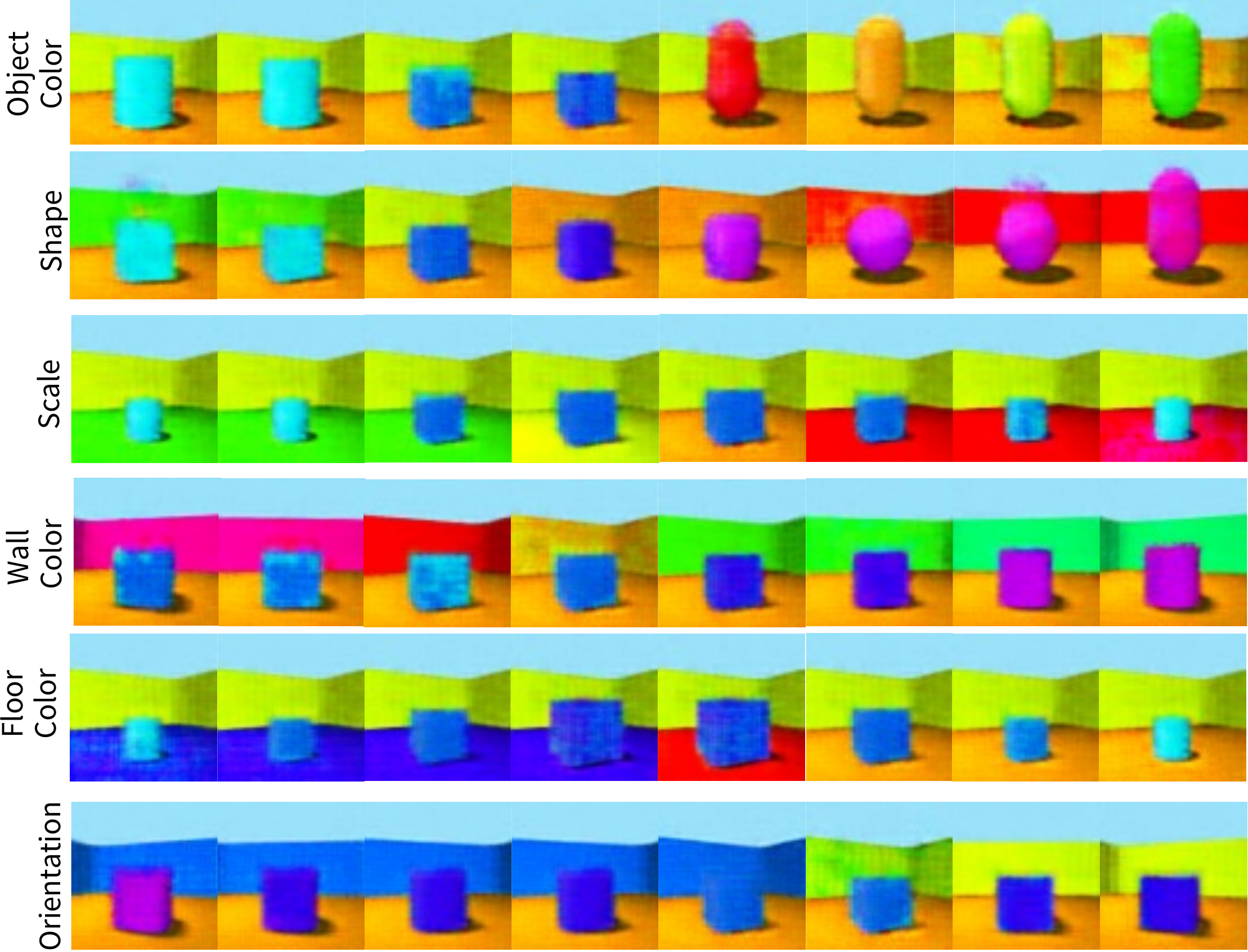}\\
(a) DiCoMoGAN & (b) $\beta$-VAE-ODE & (c) $\beta$-VAE
\end{tabular}}
\caption{\textbf{Qualitative comparisons of latent traversals on 3D Shapes~\cite{3dshapes18}.} Left: RepNet of DiCoMoGAN, Middle: $\beta$-VAE-ODE, Right: $\beta$-VAE.}
\label{fig:disentanglement}
\end{figure}

\begin{figure}[!h]
    \centering
    \includegraphics[width=0.95\linewidth]{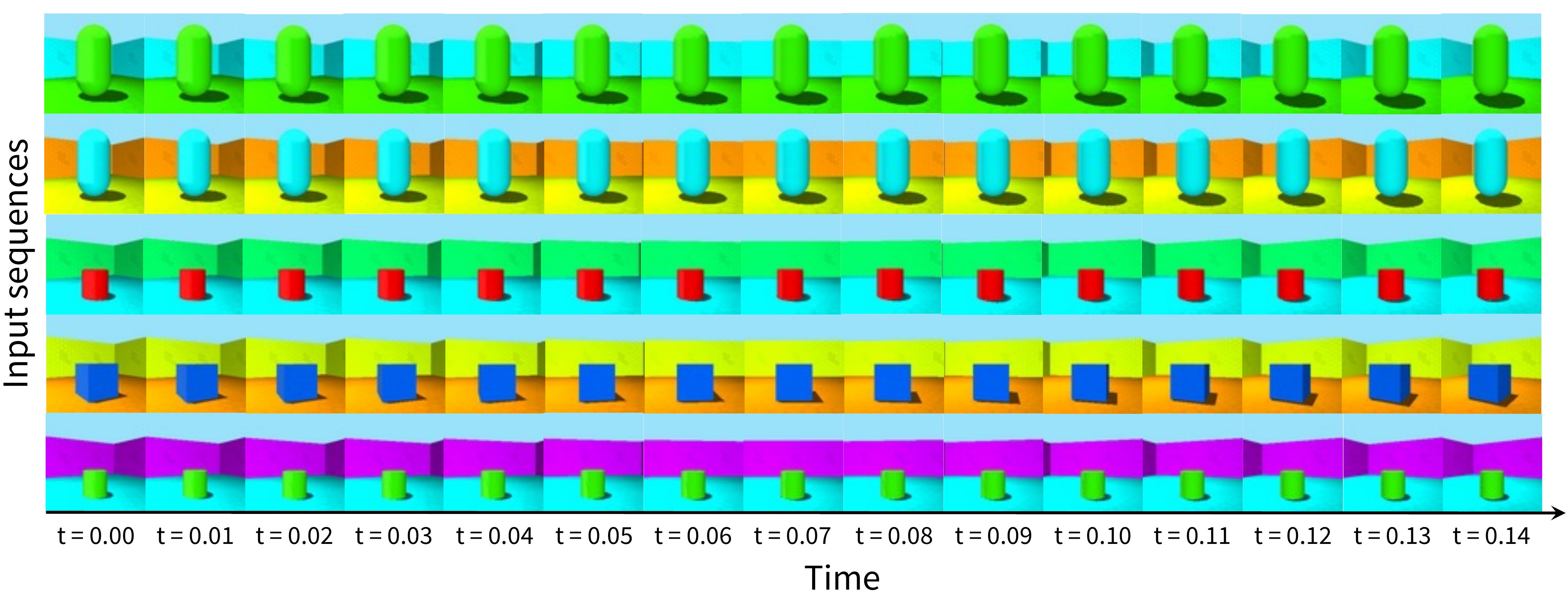}
    \includegraphics[width=0.95\linewidth]{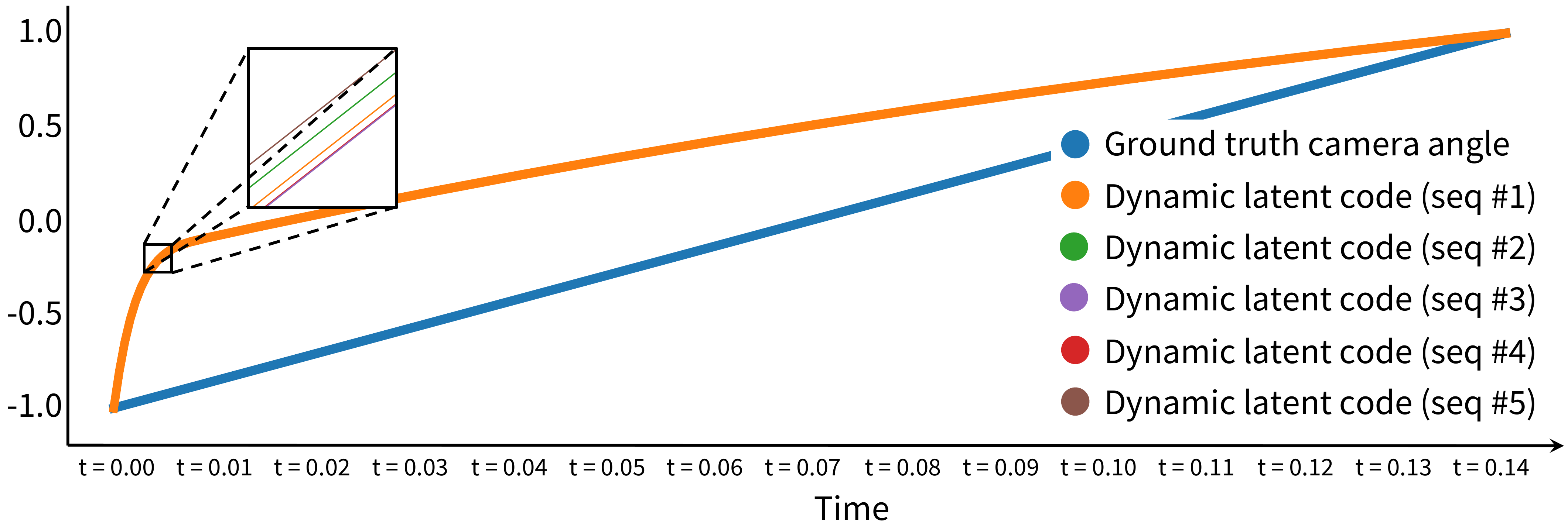}
    \caption{\textbf{Trajectories of the dynamic latent code for sample sequences from the 3D Shapes dataset~\cite{3dshapes18}.} As compared to the ground truth trajectory of the camera angle, the latent ODE component of RepNet overshoots during the interval between the first and second observations, but then the trajectories follow the same linear regime, which is consistent with the constant velocity of of the camera. As shown in the zoomed region, the differences between the sequences are negligible.}
    \label{fig:3DShapes-ODE-trajectory}
\end{figure}

\paragraph{Trajectories of Dynamic Latent Codes} 
Our proposed DiCoMoGAN model utilizes latent ODEs in the design of its RepNet to learn temporal dynamics modeling the motion observed in the observations. To illustrate this, in Fig.~\ref{fig:3DShapes-ODE-trajectory}, we show the (continuous-time) trajectories of the dynamic latent code ($\z^{\mathrm{dyn}}$) calculated by ordinary differential equations (ODEs) for some sample sequences from 3D Shapes dataset. Please remind that in our experiments on 3D Shapes, we reserve a single latent dimension for the dynamic component, which we expect to model the camera motion -- the only temporally varying factor of variation used in the construction of this synthetic dataset. Although the predicted trajectory does not perfectly fit to the ground truth trajectory, shortly after the beginning it follows a linear path which is consistent with constant angular velocity of the camera motion.

\begin{figure}[!h]
    \centering
    \includegraphics[width=\linewidth]{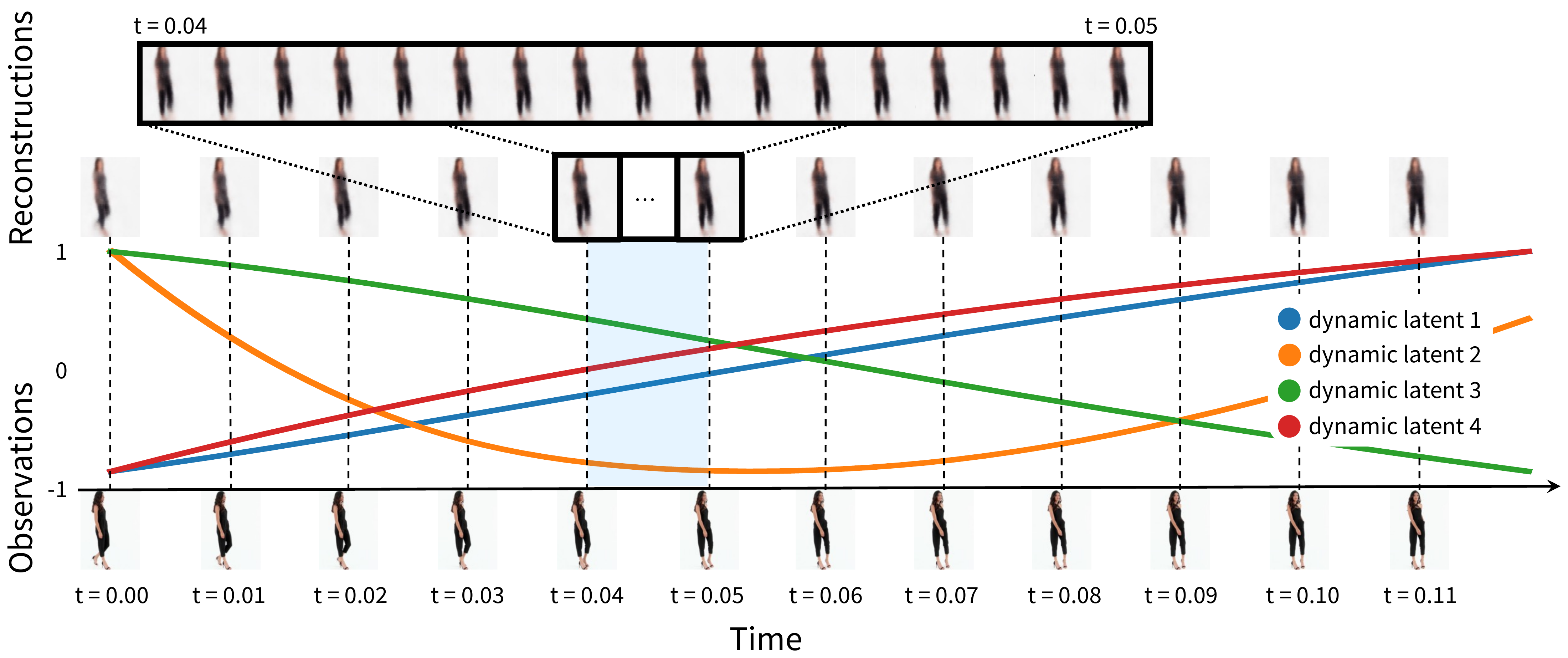}
    \caption{\textbf{Trajectories of dynamic latent codes for a sequence from Fashion Videos dataset.} The latent ODE component of RepNet allow us to interpolate between two observation, increasing the temporal resolution. For the sake of presentation, we normalize each dimension between -1 and 1.}
    \label{fig:fashion-ODE-trajectory}
\end{figure}

In Fig.~\ref{fig:fashion-ODE-trajectory}, this time we visualize the trajectories of the dynamic latent code for a sample sequence from our Fashion Videos dataset, for each of the four latent dimensions. As can be seen, compared to the case in 3D Shapes, here the trajectories are curved since the motion characteristics are more complex. Additionally, in this figure, we demonstrate an advantage of utilizing latent ODEs that it allows us to interpolate in-between frames over time.

\section{Additional Qualitative Results}
\label{sec:additional-qualitative results}
\begin{figure}[!h]
    \centering
    \includegraphics[width=\linewidth]{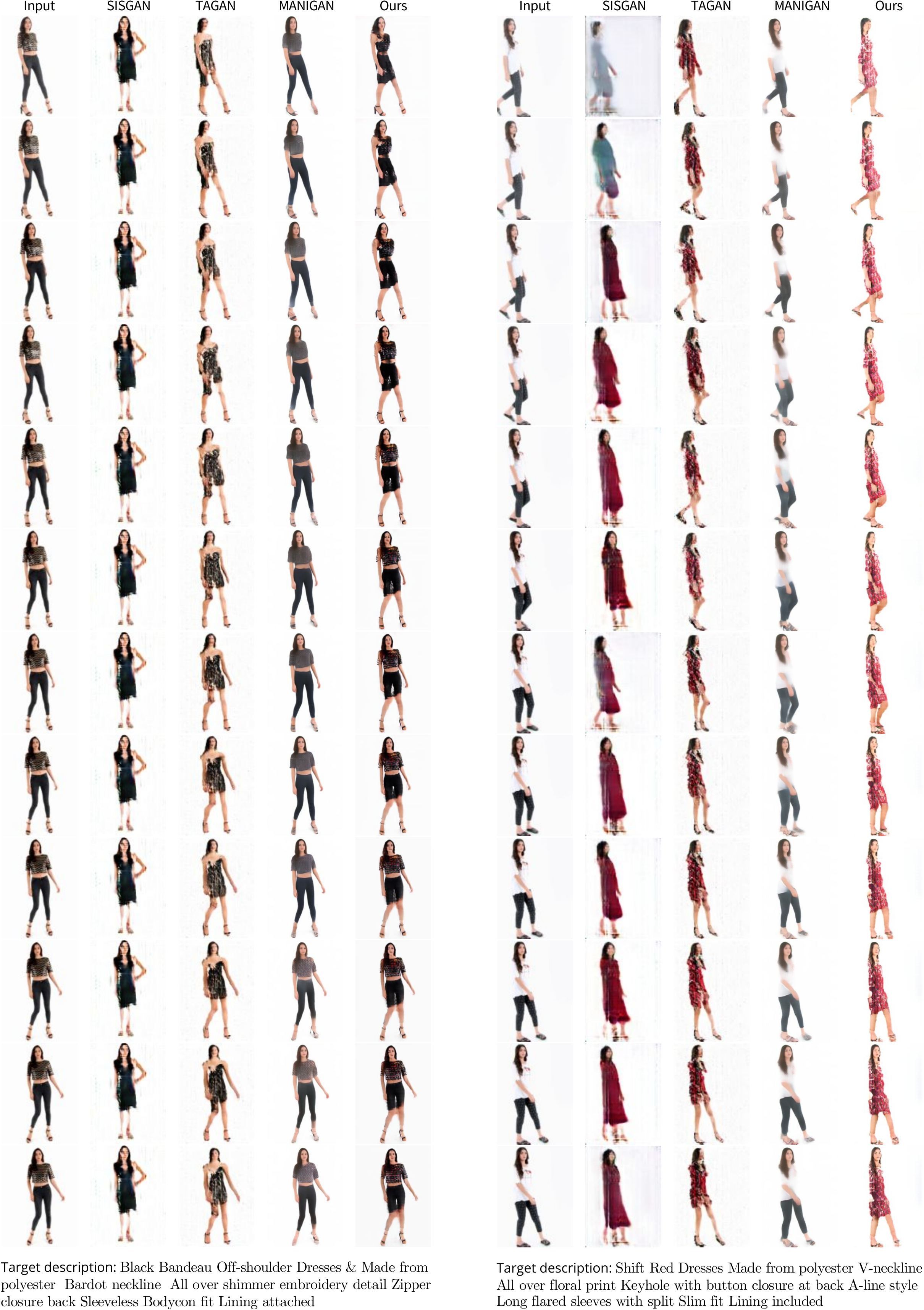}
    \caption{\textbf{Additional qualitative results.}}
    \label{fig:comparison1}
\end{figure}

\begin{figure}[!h]
    \centering
    \includegraphics[width=\linewidth]{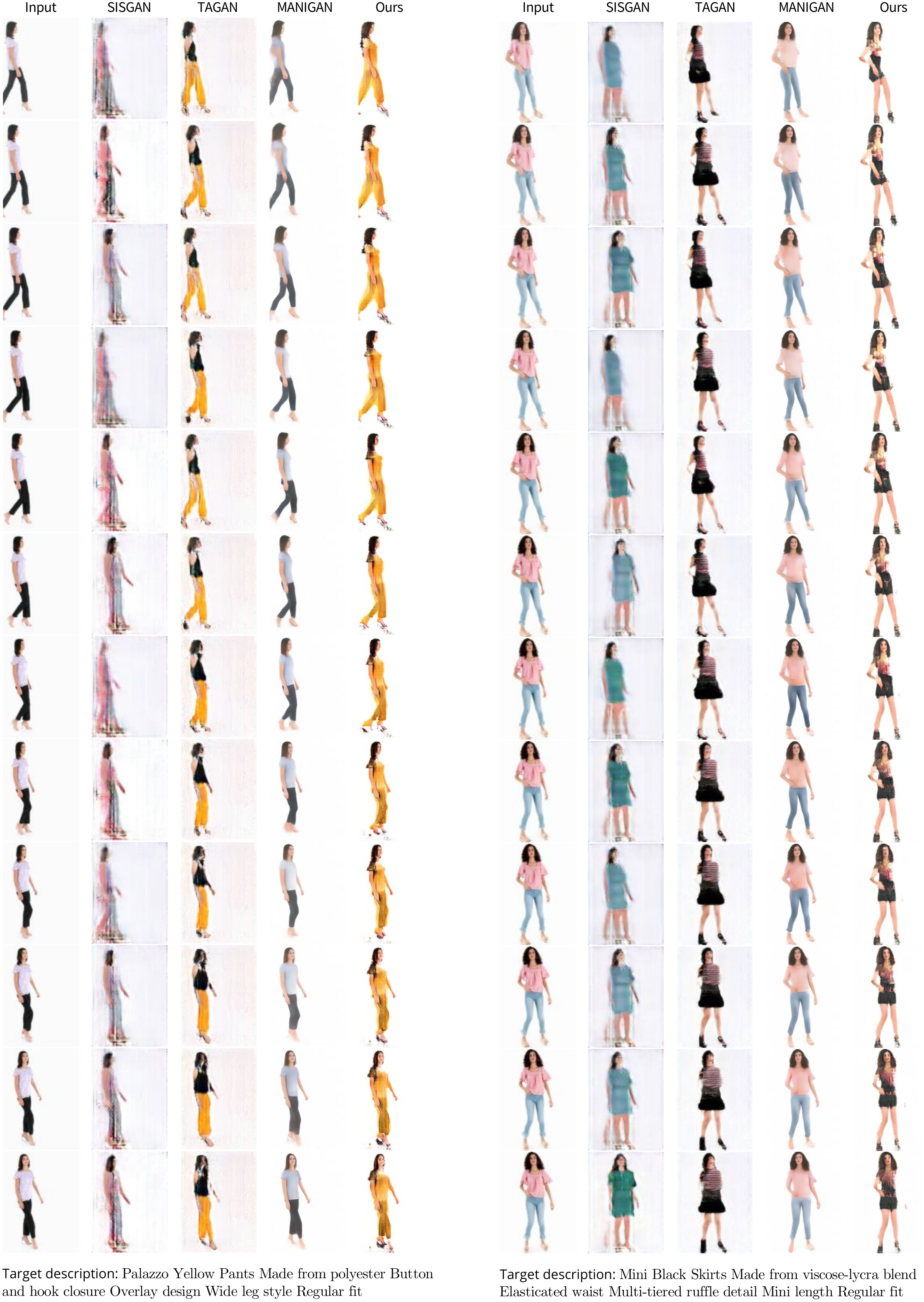}
    \caption{\textbf{Additional qualitative results.}}
    \label{fig:comparison2}
\end{figure}

Fig.~\ref{fig:comparison1} and Fig.~\ref{fig:comparison2} present %
additional visual comparisons between the proposed approach and the existing frame-based approaches, SISGAN~\cite{dong2017semantic}, TAGAN~\cite{nam2018text} and ManiGAN~\cite{li2020manigan}. Generally, our results are of better quality, containing less artifacts and structural changes semantically more relevant to the given target descriptions. As for the competing approaches, we observed than SISGAN gives the worst outcomes as it employs a simple conditioning mechanism (concatenation) to incorporate text information. The results of TAGAN are semantically more aligned with the given descriptions but their visual quality are relatively poor. ManiGAN is able to handle the changes in some attributes such as color of the garment but falls short when the target description requires some structural changes.

% \clearpage
\section{Additional Quantitative Analysis}
\label{sec:additional-quantitative-analysis}
In Table~\ref{tab:ablation_loss}, we evaluate the impact of  proposed MFMOD  and loss terms on 3dShapes and proposed Fashion Video datasets. In overall, our final loss (DiCoMoGAN) gives the best results for FID,FVD and $\text{MP}_{\text{CLIP}}$ metrics on both datasets. As for impact of MFMOD over ADAIn, proposed MFMOD provides better results for quality metrics FID, FVD and manipulation score  $\text{MP}_{\text{CLIP}}$.   

\renewcommand{\thetable}{A\arabic{table}}
\begin{table}[!h]
\centering
\caption{\textbf{Ablation study on loss functions.} 
}
\setlength{\tabcolsep}{1.2em}
\begin{tabular}{@{$\;$}c@{$\;\;\;$}l@{$\;\quad$}r@{$\;\quad$}r@{$\;\quad$}r@{$\;\quad$}r}
\toprule
&\textbf{Model} & \bf{IS} ($\uparrow$) & \bf{FID} ($\downarrow$) & \bf{FVD} ($\downarrow$) & \bf{$\text{MP}_{\text{CLIP}}$} ($\uparrow$)\\
\midrule
\parbox[t]{3mm}{\multirow{4}{*}{\rotatebox[origin=c]{90}{3D Shapes}}}&DiCoMoGAN & 2.76 & \textbf{9.08} & \textbf{69.30} & \textbf{0.26}\\
&$w/o$ MFMOD & 2.85 & 20.56 & 109.62 & \textbf{0.26}\\
&$w/o$ $\mathcal{L}_{\mathrm{L1}}$ & \textbf{3.21} & 93.15 & 811.20 & 0.24\\
&$w/o$ $\mathcal{L}_{\mathrm{unsup}}$ & 3.05 & 29.17 & 161.94 & 0.25\\
&$w/o$ $\mathcal{L}_{\mathrm{L1}}$ + $\mathcal{L}_{\mathrm{unsup}}$ & 2.68 & 212.12 & 1043.35 & 0.21\\
\bottomrule
\end{tabular}
\setlength{\tabcolsep}{1.2em}
\begin{tabular}{@{$\;$}c@{$\;\;\;$}l@{$\;\quad$}r@{$\;\quad$}r@{$\;\quad$}r@{$\;\quad$}r}
\toprule
&\textbf{Model} & \bf{IS} ($\uparrow$) & \bf{FID} ($\downarrow$) & \bf{FVD} ($\downarrow$) & \bf{$\text{MP}_{\text{CLIP}}$} ($\uparrow$)\\
\midrule
\parbox[t]{3mm}{\multirow{4}{*}{\rotatebox[origin=c]{90}{Fashion}}} & DiCoMoGAN & 2.96 & \textbf{15.34} & \textbf{53.75} & \textbf{0.25}\\
&$w/o$ MFMOD & \textbf{3.15} & 25.99 & 417.20 & 0.23\\
&$w/o$ $\mathcal{L}_{\mathrm{L1}}$ & 3.08 & 22.56 & 171.05  & 0.22\\
&$w/o$ $\mathcal{L}_{\mathrm{unsup}}$ & 3.05 & 16.78 & 76.44 & 0.23\\
&$w/o$ $\mathcal{L}_{\mathrm{L1}}$ + $\mathcal{L}_{\mathrm{unsup}}$ & 3.00 & 24.43 & 221.57 & 0.22\\
\bottomrule
\end{tabular}
\vspace{-3mm}
\label{tab:ablation_loss}
\end{table}
\section{Refinement Network}
\label{sec:refinement}
Our proposed text-based neural video manipulation approach provides qualitatively and quantitatively better results than all the existing text-based image manipulation approaches on video datasets. That said, there is still some room for improvement when fine-scale details or identity information are considered. Although our main aim in our work is to explore disentangling content and motion for obtatining better editing, we also demonstrated that finer detail can be recovered by using a local refinement network as an additional step. 

\begin{figure}[!h]
    \centering
    \includegraphics[width=0.9\linewidth]{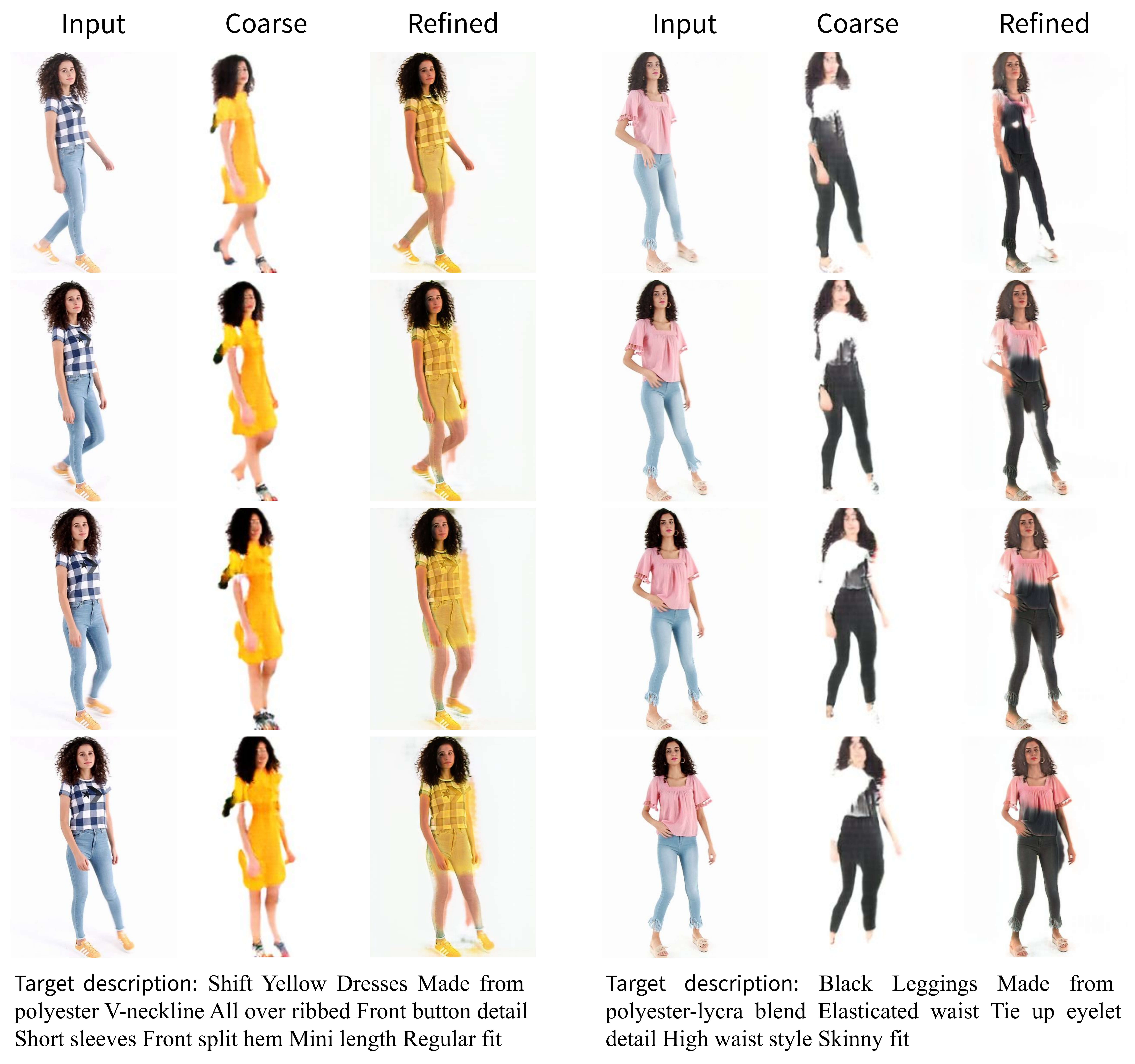}
    \caption{\textbf{Refinement Results.} Coarse manipulation results of DiCoMoGAN can be further improved by employing a 	subsequent refinement network.}
    \label{fig:refinement}
\end{figure}

Our refinement network takes the results of our DiCoMoGAN model and consists of an additional encoder, MFMOD blocks and decoder layers defined for higher resolutions. We first update the parameters of the refinement network for $10$ epochs. Then, all of our network structure containing coarse DiCoMoGAN and refinement modules is trained in an end-to-end manner for $50$ epochs. In Figure~\ref{fig:refinement}, we present sample results obtained by the aforementioned refinement process. As can be seen, the qualitative of our proposed video manipulation method can be enhanced by employing a simple multi-scale approach. 
% \newpage
\section{Comparison against GAN-inversion based editing} 
\label{sec:StyleCLIP}
\begin{figure}[!h]
\centering
\includegraphics[width=0.9\linewidth]{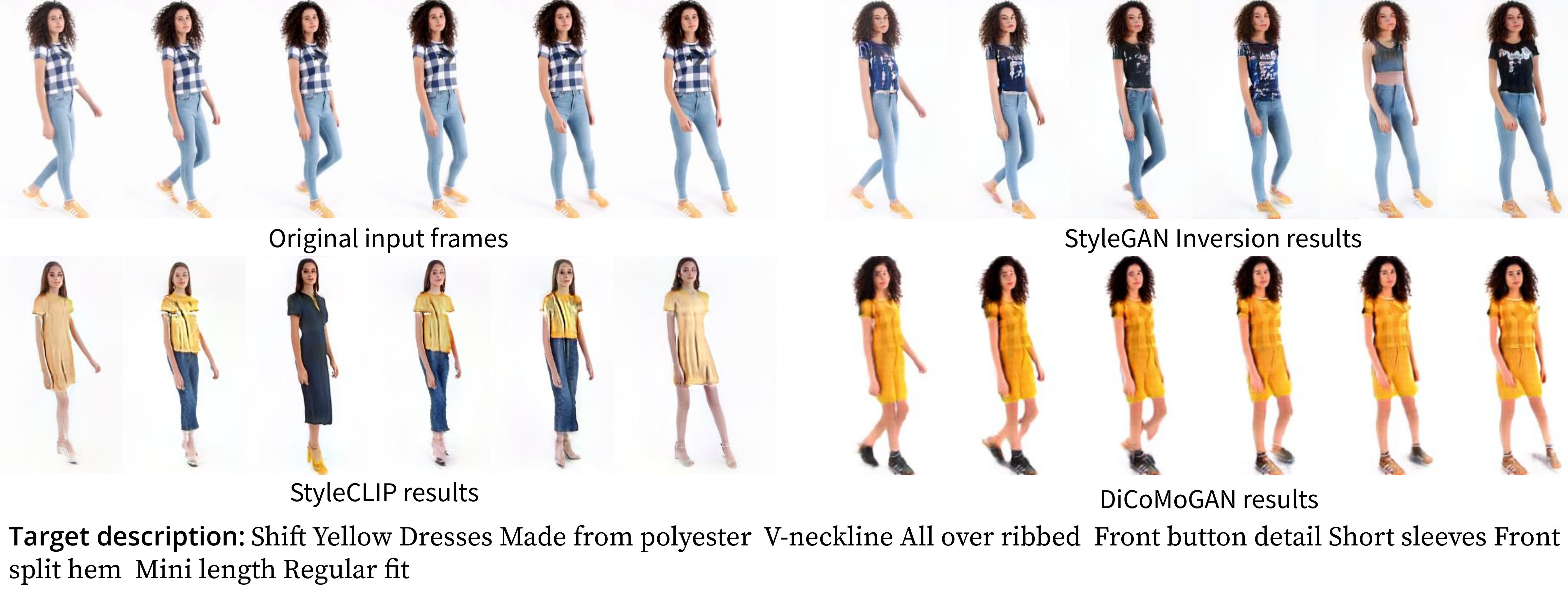}\\$\quad$\\
\includegraphics[width=0.9\linewidth]{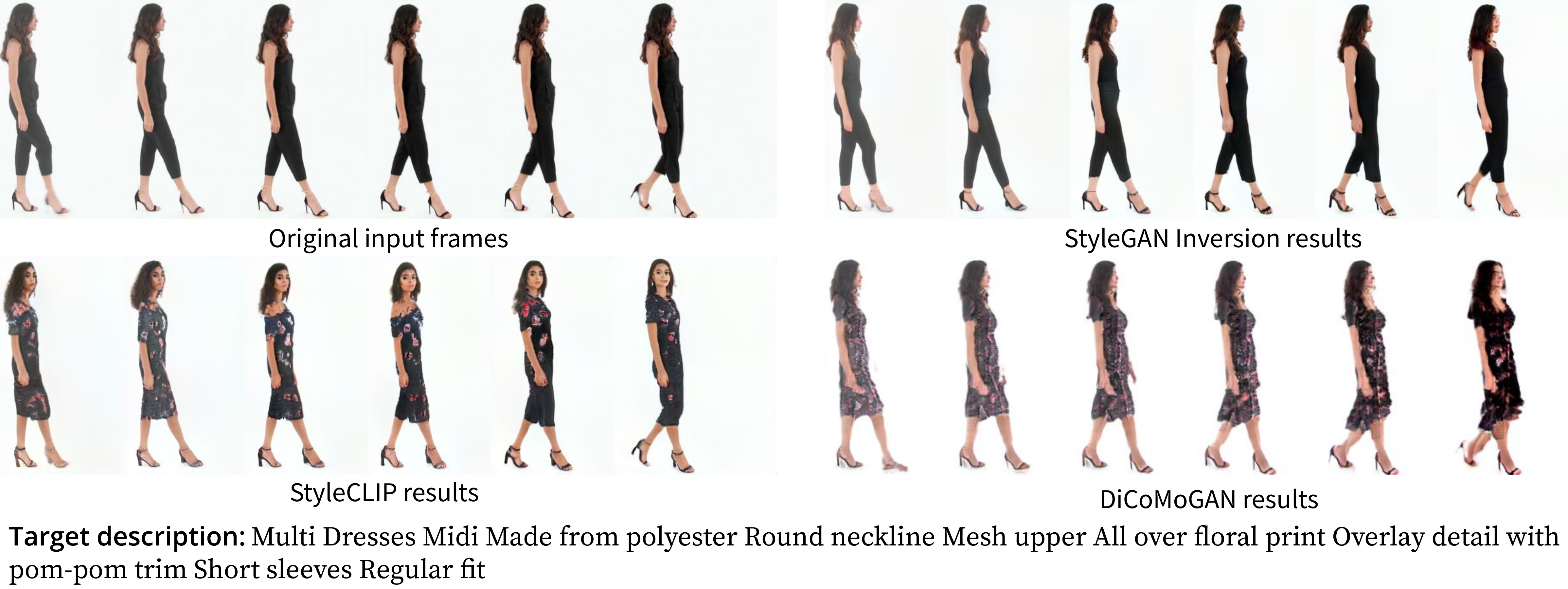}\\$\quad$\\
\includegraphics[width=0.9\linewidth]{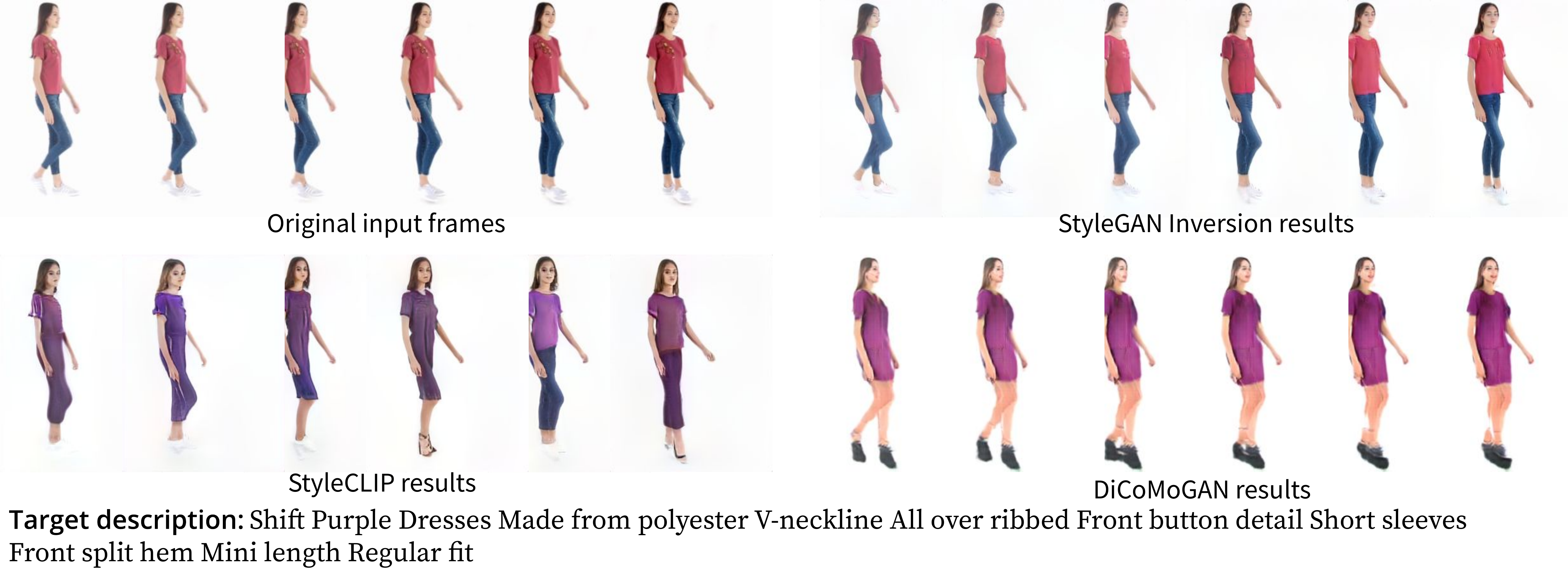}
\caption{\textbf{Comparison of DiCoMoGAN against StyleCLIP}.}
\label{fig:StyleGAN}
\end{figure}

Recently, GAN-inversion based~methods have shown a fresh perspective to uncover the power of StyleGAN models for image editing. These works, however, focus on specific domains such as aligned faces since the style-based generators do not work well on less structured data like full body images. 
For comparison, we trained a StyleGAN2 model~\cite{Karras2019stylegan2} on our Fashion Videos dataset. Fig.~\ref{fig:StyleGAN} shows some results~obtained with the StyleCLIP~\cite{Patashnik2021StyleCLIP} model, which performs text-guided edits on the learned GAN space by individually optimizing latent codes for the given description based on CLIP similarity~\cite{radford2021learning}. We observe that while manipulating the images, it fails to preserve the identity, pose and temporal continuity. Moreover, running time of StyleCLIP is significantly longer and the optimization process needs to be repeated for each new target description. Hence, we decided not to include this approach in our experimental analysis.

\end{document}